\documentclass[10pt,twocolumn,letterpaper]{article}

\usepackage{cvpr}              % To produce the CAMERA-READY version
\usepackage{graphicx}
\usepackage{amsmath}
\usepackage{amssymb}
\usepackage{booktabs}

\usepackage{hyperref}
\usepackage{amsmath}
\usepackage{comment}
\usepackage{multirow,bigdelim}
\usepackage{lipsum}
\usepackage{array}
\usepackage{wrapfig}

\usepackage{adjustbox}
\usepackage{overpic}
\usepackage{makecell}
\usepackage[normalem]{ulem}
\usepackage{blindtext}
\usepackage{xcolor}
\usepackage{soul}
\usepackage{cleveref}
\usepackage{amsfonts}
\usepackage[linesnumbered, boxed, ruled]{algorithm2e}
\newcolumntype{C}[1]{>{\centering\let\newline\\\arraybackslash\hspace{0pt}}m{#1}}

\newif\ifdraft
\drafttrue
\ifdraft
\definecolor{darkg}{rgb}{0,0.4,0}
\newcommand{\nrc}[1]{{\color{red}[\textbf{Nataniel:} #1]}}
\newcommand{\mrc}[1]{{\color{purple}[\textbf{Miki:} #1]}}
\newcommand{\vjc}[1]{{\color{blue}[\textbf{Varun:} #1]}}
\newcommand{\kac}[1]{{\color{teal}[\textbf{Kfir:} #1]}}
\newcommand{\yzc}[1]{{\color{violet}[\textbf{Yuanzhen:} #1]}}
\newcommand{\ypc}[1]{{\color{darkg}[\textbf{Yael:} #1]}}

\newcommand{\drop}[1]{}

\else
\newcommand{\nrc}[1]{}
\newcommand{\mrc}[1]{}
\newcommand{\vjc}[1]{}
\newcommand{\kac}[1]{}
\newcommand{\yzc}[1]{}
\newcommand{\ypc}[1]{}

\fi

\makeatletter
\DeclareRobustCommand\onedot{\futurelet\@let@token\@onedot}
\def\@onedot{\ifx\@let@token.\else.\null\fi\xspace}

\def\etal{\emph{et al}\onedot}
\makeatother
\usepackage[bottom]{footmisc}
\usepackage{arydshln}

\usepackage{enumitem}

\makeatletter
\def\blfootnote{\xdef\@thefnmark{}\@footnotetext}
\makeatother

\newif\ifwatermark
\watermarktrue
\draftfalse

\def\cvprPaperID{578} % *** Enter the CVPR Paper ID here
\def\confName{CVPR}
\def\confYear{2023}

\usepackage{amsmath,amsfonts,bm}
\usepackage{mathtools}

\def\eqref#1{equation~\ref{#1}}
\def\1{\bm{1}}

\DeclareMathAlphabet{\mathsfit}{\encodingdefault}{\sfdefault}{m}{sl}
\SetMathAlphabet{\mathsfit}{bold}{\encodingdefault}{\sfdefault}{bx}{n}

\newcommand{\E}{\mathbb{E}}

\newcommand{\defeq}{\coloneqq}

\newcommand{\Eb}[2]{\E_{#1}\!\left[#2\right]}

\newcommand{\bI}{\mathbf{I}}

\newcommand{\bzero}{\mathbf{0}}

\newcommand{\bc}{\mathbf{c}}

\newcommand{\bx}{\mathbf{x}}

\newcommand{\bz}{\mathbf{z}}
\newcommand{\bxh}{\hat{\mathbf{x}}}

\newcommand{\bepsilon}{{\boldsymbol{\epsilon}}}

\newcommand{\bP}{\mathbf{P}}

\newcommand{\bVh}{\hat{\mathbf{V}}}

\begin{document}

\title{\vspace{-.25in}DreamBooth: Fine Tuning Text-to-Image Diffusion Models \\ for Subject-Driven Generation\vspace{-.1in}}

\author{
  Nataniel Ruiz$^{*,1,2}$ ~~~~~~~~~~~~~~ Yuanzhen Li$^{1}$ ~~~~~~~~~~~~~~~ Varun Jampani$^{1}$ \\ ~~ Yael Pritch$^{1}$ ~~~~~~~~~~~~~~~ Michael Rubinstein$^{1}$ ~~~~~~~~~~ Kfir Aberman$^{1}$\\
  $^1$ Google Research ~~~ $^2$  Boston University\\
}

\twocolumn[{
\renewcommand\twocolumn[1][]{#1}
\maketitle
\begin{center}
    \centering
    \vspace*{-.8cm}
    \includegraphics[width=\textwidth]{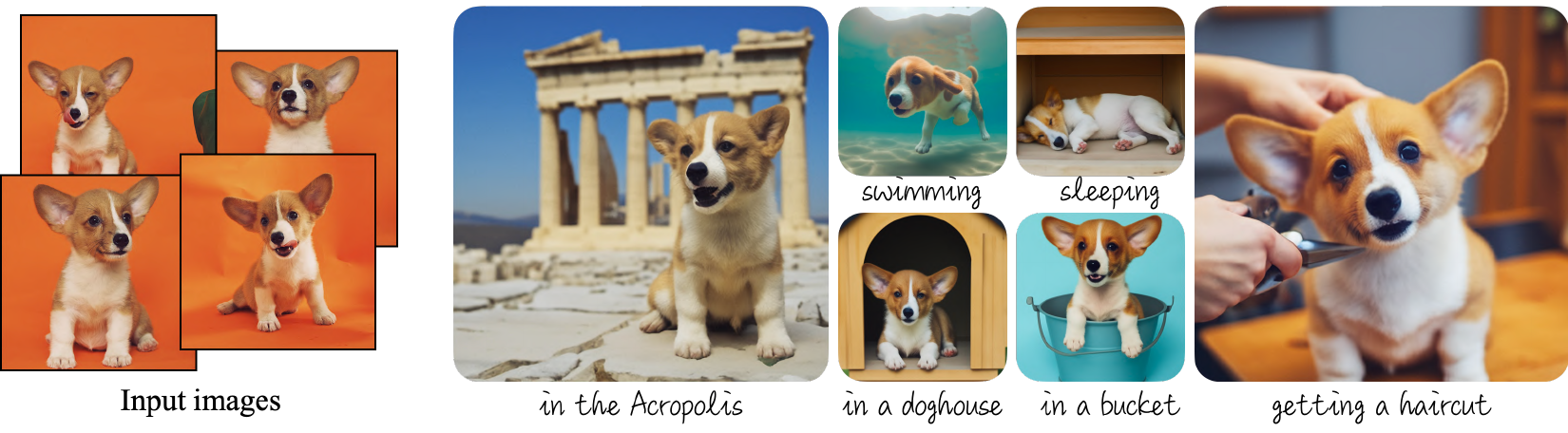}
    \vspace*{-.6cm}
    \captionof{figure}{With just a few images (typically 3-5) of a subject (left), \emph{DreamBooth}---our AI-powered photo booth---can generate a myriad of images of the subject in different contexts (right), using the guidance of a text prompt. The results exhibit natural interactions with the environment, as well as novel articulations and variation in lighting conditions, all while maintaining high fidelity to the key visual features of the subject.}
\label{fig:teaser}
\end{center}
}]

% \vspace{-.5in}
\blfootnote{*This research was performed while Nataniel Ruiz was at Google.}
\begin{abstract}
Large text-to-image models achieved a remarkable leap in the evolution of AI, enabling high-quality and diverse synthesis of images from a given text prompt. However, these models lack the ability to mimic the appearance of subjects in a given reference set and synthesize novel renditions of them in different contexts. In this work, we present a new approach for ``personalization'' of text-to-image diffusion models. Given as input just a few images of a subject, we fine-tune a pretrained text-to-image model such that it learns to bind a unique identifier with that specific subject. Once the subject is embedded in the output domain of the model, the unique identifier can be used to synthesize novel photorealistic images of the subject contextualized in different scenes. By leveraging the semantic prior embedded in the model with a new autogenous class-specific prior preservation loss, our technique enables synthesizing the subject in diverse scenes, poses, views and lighting conditions that do not appear in the reference images. We apply our technique to several previously-unassailable tasks, including subject recontextualization, text-guided view synthesis, and artistic rendering, all while preserving the subject's key features. We also provide a new dataset and evaluation protocol for this new task of subject-driven generation. Project page:~\small{\url{https://dreambooth.github.io/}}
\end{abstract}

\section{Introduction}

Can you imagine your own dog traveling around the world, or your favorite bag displayed in the most exclusive showroom in Paris? What about your parrot being the main character of an illustrated storybook?
Rendering such imaginary scenes is a challenging task that requires synthesizing instances of specific subjects (e.g., objects, animals) in new contexts such that they naturally and seamlessly blend into the scene.

\begin{figure}[t]
\centering
\includegraphics[clip,width=.9\columnwidth]{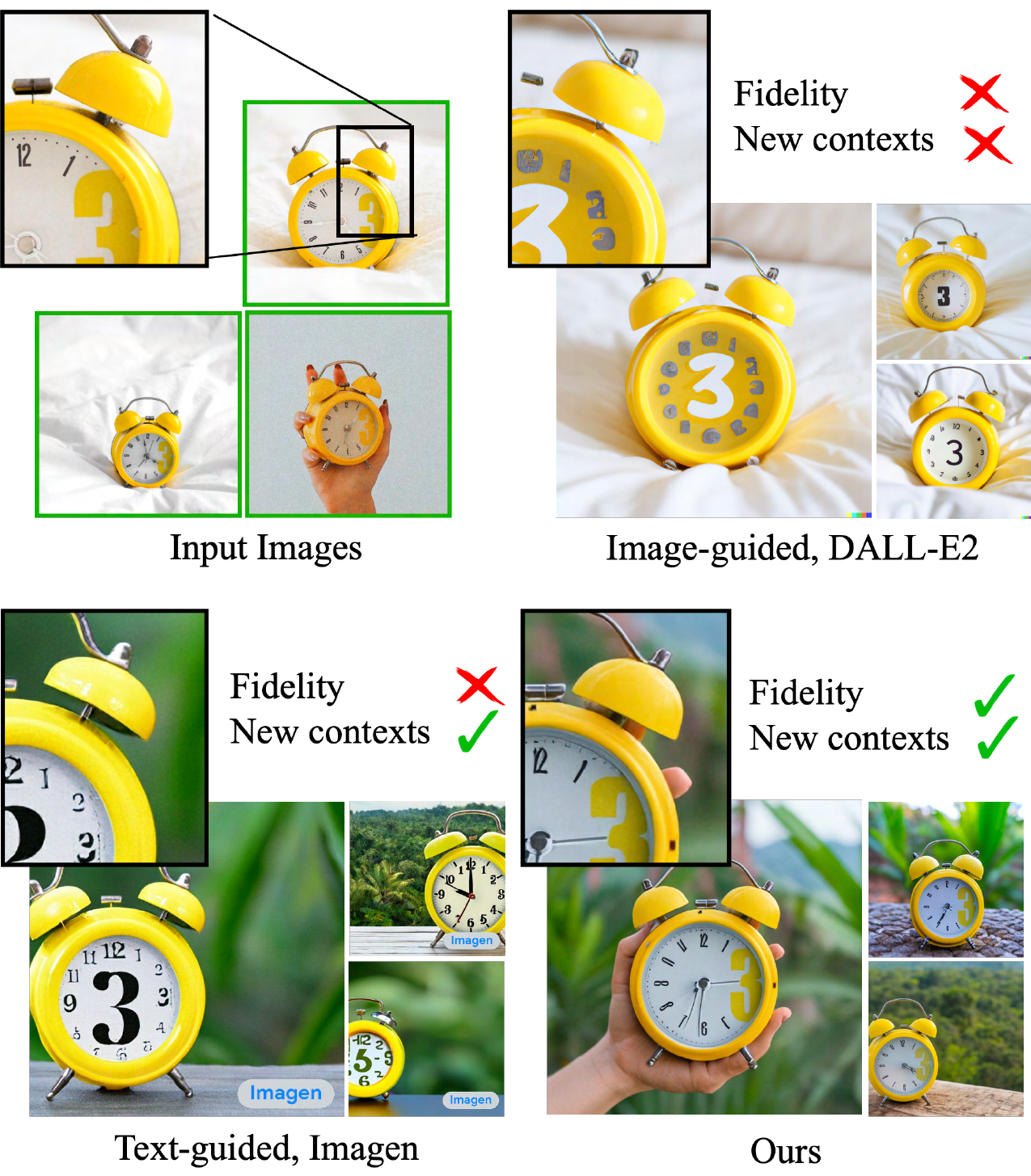}
\caption{{\bf Subject-driven generation.} Given a particular clock (left), it is hard to generate it while maintaining high fidelity to its key visual features (second and third columns showing DALL-E2~\cite{ramesh2022hierarchical} image-guided generation and Imagen~\cite{saharia2022photorealistic} text-guided generation; text prompt used for Imagen: \textit{``retro style yellow alarm clock with a white clock face and a yellow number three on the right part of the clock face in the jungle"}). Our approach (right) can synthesize the clock with high fidelity and in new contexts (text prompt: \textit{``a [V] clock in the jungle"}).}
\label{fig:explanatory_figure}
\end{figure}

Recently developed large text-to-image models have shown unprecedented capabilities, by enabling high-quality and diverse synthesis of images based on a text prompt written in natural language \cite{saharia2022photorealistic,ramesh2022hierarchical}. One of the main advantages of such models is the strong semantic prior learned from a large collection of image-caption pairs. Such a prior learns, for instance, to bind the word ``dog" with various instances of dogs that can appear in different poses and contexts in an image.  
While the synthesis capabilities of these models are unprecedented, they lack the ability to mimic the appearance of subjects in a given reference set, and synthesize novel renditions of the \emph{same subjects} in different contexts. The main reason is that the expressiveness of their output domain is limited;  even the most detailed textual description of an object may yield instances with different appearances. Furthermore, even models whose text embedding lies in a shared language-vision space \cite{radford2021learning} cannot accurately reconstruct the appearance of given subjects but only create variations of the image content (Figure~\ref{fig:explanatory_figure}). 

In this work, we present a new approach for ``personalization'' of text-to-image diffusion models (adapting them to user-specific image generation needs). Our goal is to expand the language-vision dictionary of the model such that it binds new words with specific subjects the user wants to generate. Once the new dictionary is embedded in the model, it can use these words to synthesize novel photorealistic images of the subject, contextualized in different scenes, while preserving their key identifying features. The effect is akin to a ``magic photo booth''---once a few images of the subject are taken, the booth generates photos of the subject in different conditions and scenes, as guided by simple and intuitive text prompts (Figure~\ref{fig:teaser}).

More formally, given a few images of a subject ($\sim$3-5), our objective is to implant the subject into the output domain of the model such that it can be synthesized with a \emph{unique identifier}. To that end, we propose a technique to represent a given subject with rare token identifiers and fine-tune a pre-trained, diffusion-based text-to-image framework. 

We fine-tune the text-to-image model with the input images and text prompts containing a unique identifier followed by the class name of the subject (e.g., ``A [V] dog”). The latter enables the model to use its prior knowledge on the subject class while the class-specific instance is bound with the unique identifier. In order to prevent \textit{language drift}~\cite{Lee2019CounteringLD,lu2020countering} that causes the model to associate the class name (e.g., ``dog") with the specific instance, we propose an \textit{autogenous, class-specific prior preservation loss}, which leverages the semantic prior on the class that is embedded in the model, and encourages it to generate diverse instances of the same class as our subject.

We apply our approach to a myriad of text-based image generation applications including recontextualization of subjects, modification of their properties, original art renditions, and more, paving the way to a new stream of previously unassailable tasks. We highlight the contribution of each component in our method via ablation studies, and compare with alternative baselines and related work. We also conduct a user study to evaluate subject and prompt fidelity in our synthesized images, compared to alternative approaches.

To the best of our knowledge, ours is the first technique that tackles this new challenging problem of subject-driven generation, allowing users, from just a few casually captured images of a subject, synthesize novel renditions of the subject in different contexts while maintaining its distinctive features.

To evaluate this new task, we also construct a new dataset that contains various subjects captured in different contexts, and propose a new evaluation protocol that measures the subject fidelity and prompt fidelity of the generated results. We make our dataset and evaluation protocol publicly available on the project webpage.

\section{Related work}

\textbf{Image Composition. }
Image composition techniques \cite{wu2019gp,cong2020dovenet,lin2018st} aim to clone a given subject into a new background such that the subject melds into the scene. To consider composition in novel poses, one may apply 3D reconstruction techniques \cite{mildenhall2020nerf,Barron_2021_ICCV,boss2022samurai,verbin2021refnerf,poole2022dreamfusion} which usually works on rigid objects and require a larger number of views. Some drawbacks include scene integration (lighting, shadows, contact) and the inability to generate novel scenes. In contrast, our approach enable generation of subjects in novel poses and new contexts.

\textbf{Text-to-Image Editing and Synthesis. }
Text-driven image manipulation has recently achieved significant progress using GANs
~\cite{goodfellow2014generative,brock2018large,karras2021alias,karras2019style,karras2020analyzing} combined with image-text representations such as CLIP~\cite{radford2021learning}, yielding realistic manipulations using text~\cite{patashnik2021styleclip,gal2021stylegan, xia2021tedigan, abdal2021clip2stylegan,bau2021paint,mokady2022self}.
These methods work well on structured scenarios (e.g. human face editing) and can struggle over diverse datasets where subjects are varied.
Crowson et al. \cite{crowson2022vqgan} use VQ-GAN \cite{esser2021taming} and train over more diverse data to alleviate this concern.
Other works~\cite{avrahami2022blended, kim2022diffusionclip} exploit the recent diffusion models \cite{ho2020denoising,sohl2015deep,song2019generative,ho2020denoising,song2020denoising,rombach2021highresolution,nichol2021improved,song2020improved,palette,saharia2021image}, which achieve state-of-the-art generation quality over highly diverse datasets, often surpassing GANs~\cite{dhariwal2021diffusion}.
While most works that require only text are limited to global editing \cite{crowson2022vqgan, kwon2021clipstyler}, 
Bar-Tal et al.~\cite{bar2022text2live} proposed a text-based localized editing technique without using masks, showing impressive results. While most of these editing approaches allow modification of global properties or local editing of a given image, none enables generating novel renditions of a given subject in new contexts.

There also exists work on text-to-image synthesis \cite{ding2021cogview,hinz2020semantic,tao2020df, li2019controllable,li2019object,qiao2019learn,qiao2019mirrorgan,ramesh2021zero,zhang2018photographic,crowson2022vqgan, gafni2022make, rombach2021highresolution,jain2021dreamfields}. Recent large text-to-image models such as Imagen~\cite{saharia2022photorealistic}, DALL-E2~\cite{ramesh2022hierarchical}, Parti~\cite{yu2022scaling}, CogView2~\cite{ding2022cogview2} and Stable Diffusion~\cite{rombach2021highresolution} demonstrated unprecedented semantic generation. These models do not provide fine-grained control over a generated image and use text guidance only.
Specifically, it is challenging or impossible to preserve the identity of a subject consistently across synthesized images.

\textbf{Controllable Generative Models. }
There are various approaches to control generative models, where some of them might prove to be viable directions for subject-driven prompt-guided image synthesis.
Liu et al.~\cite{liu2019more} propose a diffusion-based technique allowing for image variations guided by reference image or text.
To overcome subject modification, several works~\cite{nichol2021glide, avrahami2022blendedlatent} assume a user-provided mask to restrict the modified area. Inversion~\cite{choi2021ilvr,dhariwal2021diffusion,ramesh2022hierarchical} can be used to preserve a subject while modifying context. Prompt-to-prompt~\cite{hertz2022prompt} allows for local and global editing without an input mask. These methods fall short of identity-preserving novel sample generation of a subject.

In the context of GANs, Pivotal Tuning~\cite{roich2021pivotal} allows for real image editing by finetuning the model with an inverted latent code anchor, and Nitzan et al.~\cite{nitzan2022mystyle} extended this work to GAN finetuning on faces to train a personalized prior, which requires around 100 images and are limited to the face domain. Casanova et al.~\cite{casanova2021instance} propose an instance conditioned GAN that can generate variations of an instance, although it can struggle with unique subjects and does not preserve all subject details. 
 
Finally, the concurrent work of Gal \etal\cite{gal2022image} proposes a method to represent visual concepts, like an object or a style, through new tokens in the embedding space of a frozen text-to-image model, resulting in small personalized token embeddings. While this method is limited by the expressiveness of the frozen diffusion model, our fine-tuning approach enables us to embed the subject within the model's output domain, resulting in the generation of novel images of the subject which preserve its key visual features.

\section{Method}
\label{sec:method}

Given only a few (typically 3-5) casually captured images of a specific subject, without any textual description, our objective is to generate new images of the subject with high detail fidelity and with variations guided by text prompts. Example variations include changing the subject location, changing subject properties such as color or shape, modifying the subject's pose, viewpoint, and other semantic modifications. We do not impose any restrictions on input image capture settings and the subject image can have varying contexts. We next provide some background on text-to-image diffusion models (Sec.~\ref{sec:t2i}), then present our fine-tuning technique to bind a unique identifier with a subject described in a few images (Sec.~\ref{sec:finetune}), and finally propose a class-specific prior-preservation loss that enables us to overcome language drift in our fine-tuned model (Sec.~\ref{sec:method_prior_pres}).

\subsection{Text-to-Image Diffusion Models}
\label{sec:t2i}
Diffusion models are probabilistic generative models that are trained to learn a data distribution by the gradual denoising of a variable sampled from a Gaussian distribution. Specifically, we are interested in a pre-trained text-to-image diffusion model $\hat\bx_\theta$ that, given an initial noise map $\bepsilon \sim \mathcal{N}(\bzero, \bI)$ and a conditioning vector $\bc=\Gamma(\bP)$ generated using a text encoder $\Gamma$ and a text prompt $\bP$, generates an image $\bx_{\text{gen}}=\hat\bx_\theta(\bepsilon, \bc)$. They are trained using a squared error loss to denoise a variably-noised image or latent code $\bz_t \coloneqq \alpha_t \bx + \sigma_t \bepsilon$ as follows:
\begin{equation}
\label{eq:diffusion}
    \Eb{\bx,\bc,\bepsilon,t}{w_t \|\hat\bx_\theta(\alpha_t \bx + \sigma_t \bepsilon, \bc) - \bx \|^2_2}
\end{equation}
where $\bx$ is the ground-truth image, $\bc$ is a conditioning vector (e.g., obtained from a text prompt), and $\alpha_t, \sigma_t, w_t$ are terms that control the noise schedule and sample quality, and are functions of the diffusion process time $t \sim \mathcal{U}([0, 1])$. 
A more detailed description is given in the supplementary material.  

\subsection{Personalization of Text-to-Image Models}
\label{sec:finetune}

% \begin{figure*}[t]
% \centering
% \includegraphics[trim={0 0 0 0},clip,width=\textwidth]{figures/system_fig.pdf}
% \caption{\textbf{Fine-tuning.} Given $\sim3-5$ images of a subject we fine tune a text-to-image diffusion in two steps: (a) fine tuning the low-resolution text-to-image model with the input images paired with a text prompt containing a unique identifier and the name of the class the subject belongs to (e.g., ``A [V] dog”), in parallel, we apply a class-specific prior preservation loss, which leverages the semantic prior that the model has on the class and encourages it to generate diverse instances belong to the subject's class using the class name in a text prompt (e.g., ``A dog”). (b) fine-tuning the super resolution components with pairs of low-resolution and high-resolution images taken from our input images set, which enables us to maintain high-fidelity to small details of the subject.}
% \label{fig:system_overview} 
% \end{figure*}
\begin{figure}[t]
\centering
\includegraphics[width=\columnwidth]{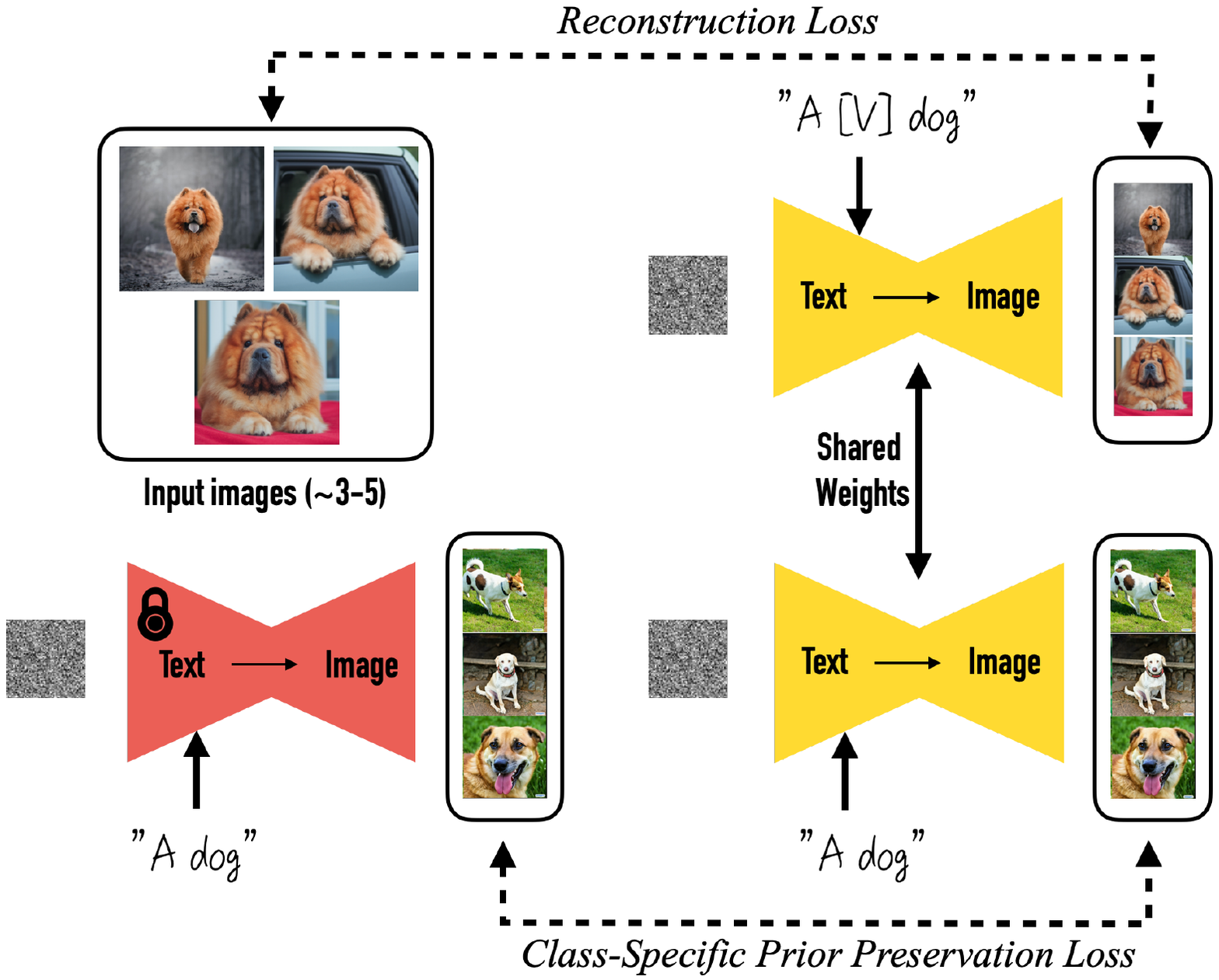}
\caption{\textbf{Fine-tuning.} Given $\sim3-5$ images of a subject we fine-tune a text-to-image diffusion model with the input images paired with a text prompt containing a unique identifier and the name of the class the subject belongs to (e.g., ``A [V] dog”), in parallel, we apply a class-specific prior preservation loss, which leverages the semantic prior that the model has on the class and encourages it to generate diverse instances belong to the subject's class using the class name in a text prompt (e.g., ``A dog”). }
\label{fig:system_overview} 
\end{figure}

Our first task is to implant the subject instance into the output domain of the model such that we can query the model for varied novel images of the subject. One natural idea is to fine-tune the model using the few-shot dataset of the subject. Careful care had to be taken when fine-tuning generative models such as GANs in a few-shot scenario as it can cause overfitting and mode-collapse - as well as not capturing the target distribution sufficiently well. There has been research on techniques to avoid these pitfalls~\cite{Robb2020FewShotAO,ojha2021few,li2020few,mo2020freeze,Wang_2020_CVPR}, although, in contrast to our work, this line of work primarily seeks to generate images that resemble the target distribution but has no requirement of subject preservation. With regards to these pitfalls, we observe the peculiar finding that, given a careful fine-tuning setup using the diffusion loss from Eq~\ref{eq:diffusion}, large text-to-image diffusion models seem to excel at integrating new information into their domain without forgetting the prior or overfitting to a small set of training images. 

\paragraph{Designing Prompts for Few-Shot Personalization}
Our goal is to ``implant" a new (\textit{unique identifier}, subject) pair into the diffusion model's ``dictionary'' .
In order to bypass the overhead of writing detailed image descriptions for a given image set we opt for a simpler approach and label all input images of the subject ``a [identifier] [class noun]", where [identifier] is a unique identifier linked to the subject and [class noun] is a coarse class descriptor of the subject (e.g. cat, dog, watch, etc.). The class descriptor can be provided by the user or obtained using a classifier. We use a class descriptor in the sentence in order to tether the prior of the class to our unique subject and find that using a wrong class descriptor, or no class descriptor increases training time and language drift while decreasing performance. In essence, we seek to leverage the model's prior of the specific class and entangle it with the embedding of our subject's unique identifier so we can leverage the visual prior to generate new poses and articulations of the subject in different contexts.

\paragraph{Rare-token Identifiers}
We generally find existing English words (e.g. ``unique'', ``special'') suboptimal since the model has to learn to disentangle them from their original meaning and to re-entangle them to reference our subject. This motivates the need for an identifier that has a weak prior in both the language model and the diffusion model. A hazardous way of doing this is to select random characters in the English language and concatenate them to generate a rare identifier (e.g. ``xxy5syt00"). In reality, the tokenizer might tokenize each letter separately, and the prior for the diffusion model is strong for these letters. We often find that these tokens incur the similar weaknesses as using common English words.
Our approach is to find rare tokens in the vocabulary, and then invert these tokens into text space, in order to minimize the probability of the identifier having a strong prior. We perform a rare-token lookup in the vocabulary and obtain a sequence of rare token identifiers $f(\bVh)$, where $f$ is a tokenizer; a function that maps character sequences to tokens and $\bVh$ is the decoded text stemming from the tokens $f(\bVh)$. The sequence can be of variable length $k$, and find that relatively short sequences of $k=\{1, ..., 3\}$ work well. Then, by inverting the vocabulary using the de-tokenizer on $f(\bVh)$ we obtain a sequence of characters that define our unique identifier $\bVh$. For Imagen, we find that using uniform random sampling of tokens that correspond to 3 or fewer Unicode characters (without spaces) and using tokens in the T5-XXL tokenizer range of $\{5000, ..., 10000\}$ works well.

\subsection{Class-specific Prior Preservation Loss}
\label{sec:method_prior_pres}
In our experience, the best results for maximum subject fidelity are achieved by fine-tuning all layers of the model. This includes fine-tuning layers that are conditioned on the text embeddings, which gives rise to the problem of \textit{language drift}. Language drift has been an observed problem in language models~\cite{Lee2019CounteringLD,lu2020countering}, where a model that is pre-trained on a large text corpus and later fine-tuned for a specific task progressively loses syntactic and semantic knowledge of the language. To the best of our knowledge, we are the first to find a similar phenomenon affecting diffusion models, where to model slowly forgets how to generate subjects of the same class as the target subject.

Another problem is the possibility of \textit{reduced output diversity}. Text-to-image diffusion models naturally posses high amounts of output diversity. When fine-tuning on a small set of images we would like to be able to generate the subject in novel viewpoints, poses and articulations. Yet, there is a risk of reducing the amount of variability in the output poses and views of the subject (e.g. snapping to the few-shot views). We observe that this is often the case, especially when the model is trained for too long.

To mitigate the two aforementioned issues, we propose an autogenous class-specific prior preservation loss that encourages diversity and counters language drift. In essence, our method is to supervise the model with its \textit{own generated samples}, in order for it to retain the prior once the few-shot fine-tuning begins. This allows it to generate diverse images of the class prior, as well as retain knowledge about the class prior that it can use in conjunction with knowledge about the subject instance. Specifically, we generate data $\bx_\text{pr}=\bxh(\bz_{t_1}, \bc_\text{pr})$ by using the ancestral sampler on the frozen pre-trained diffusion model with random initial noise $\bz_{t_1} \sim \mathcal{N}(\bzero, \bI)$ and conditioning vector $\bc_\text{pr} \defeq \Gamma(f(\text{"a [class noun]"}))$. The loss becomes:
\begin{multline}
\label{eq:prior_pres}
    \E_{\bx,\bc,\bepsilon,\bepsilon^\prime,t}[w_t \|\hat\bx_\theta(\alpha_t \bx + \sigma_t \bepsilon, \bc) - \bx \|^2_2 + \\
    \lambda w_{t^\prime} \|\hat\bx_\theta(\alpha_{t^\prime} \bx_\text{pr} + \sigma_{t^\prime} \bepsilon^\prime, \bc_\text{pr}) - \bx_\text{pr} \|^2_2],
\end{multline}
where the second term is the prior-preservation term that supervises the model with its own generated images, and $\lambda$ controls for the relative weight of this term. Figure~\ref{fig:system_overview} illustrates the model fine-tuning with the class-generated samples and prior-preservation loss. 
Despite being simple, we find this prior-preservation loss is effective in encouraging output diversity and  in overcoming language-drift. We also find that we can train the model for more iterations without risking overfitting. We find that \(\sim \) 1000 iterations with $\lambda = 1$ and learning rate $10^{-5}$ for Imagen~\cite{saharia2022photorealistic} and $5 \times 10^{-6}$ for Stable Diffusion~\cite{rombach2022high}, and with a subject dataset size of 3-5 images is enough to achieve good results. During this process, $\sim 1000$ ``a [class noun]'' samples are generated - but less can be used. The training process takes about 5 minutes on one TPUv4 for Imagen, and 5 minutes on a NVIDIA A100 for Stable Diffusion.

\section{Experiments} 
\label{sec:app}

In this section, we show experiments and applications. Our method enables a large expanse of text-guided semantic modifications of our subject instances, including recontextualization, modification of subject properties such as material and species, art rendition, and viewpoint modification. Importantly, across all of these modifications, we are able to \textbf{preserve the unique visual features that give the subject its identity and essence}. If the task is recontextualization, then the subject features are unmodified, but appearance (e.g., pose) may change. If the task is a stronger semantic modification, such as crossing between our subject and another species/object, then the key features of the subject are preserved after modification. In this section, we reference the subject's unique identifier using [V]. We include specific Imagen and Stable Diffusion implementation details in the supp. material.

\begin{figure}[h!]
\centering
\includegraphics[clip,width=\columnwidth]{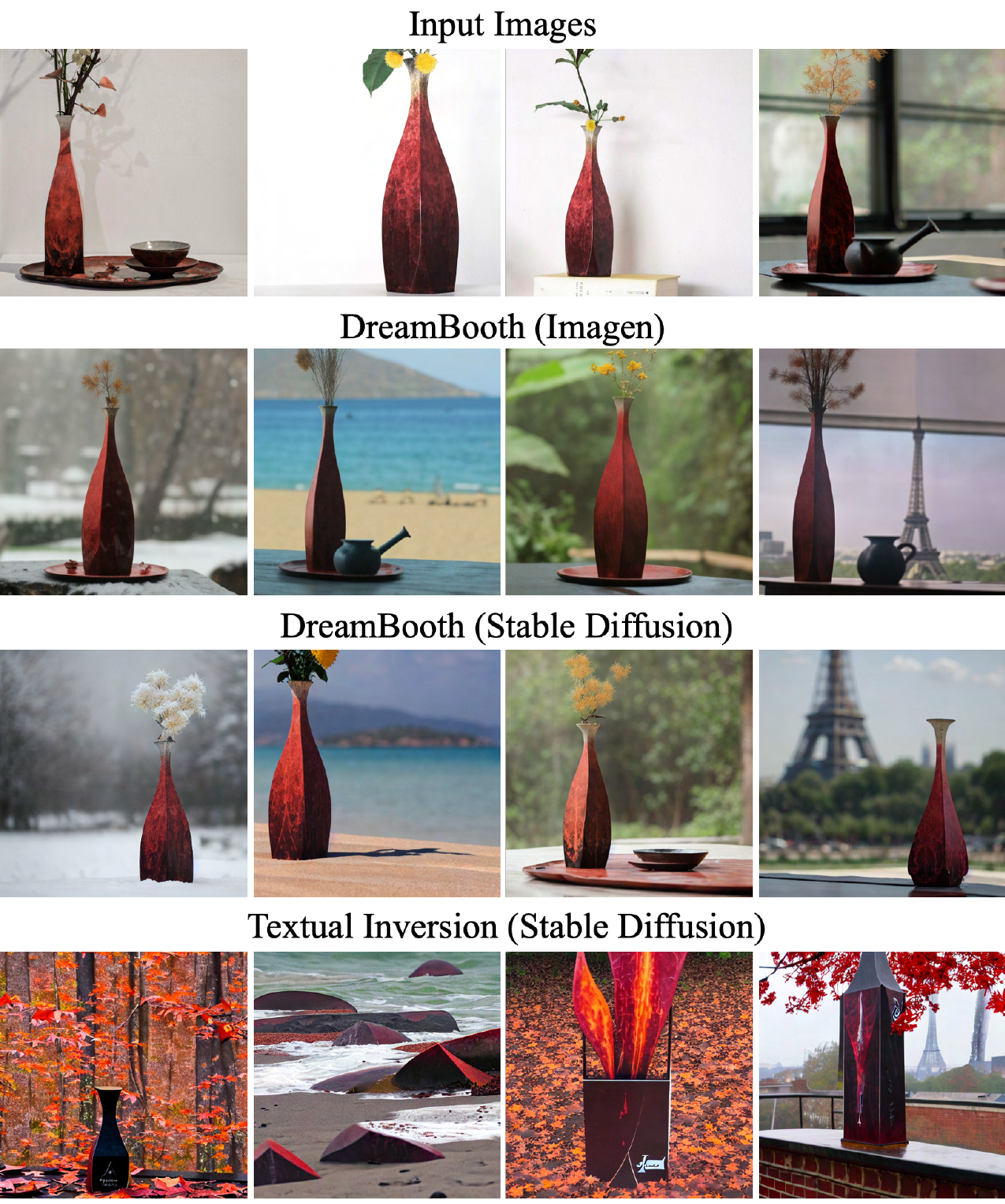}
\caption[]{
\textbf{Comparisons with Textual Inversion~\cite{gal2022image}} Given 4 input images (top row), we compare: DreamBooth Imagen (2nd row), DreamBooth Stable Diffusion (3rd row), Textual Inversion (bottom row). Output images were created with the following prompts (left to right): ``a [V] vase in the snow", ``a [V] vase on the beach", ``a [V] vase in the jungle", ``a [V] vase with the Eiffel Tower in the background".
DreamBooth is stronger in both subject and prompt fidelity.
\label{fig:comparison_gal}}
\end{figure}

\begin{figure*}[t]
\centering
\includegraphics[clip,width=1\textwidth]{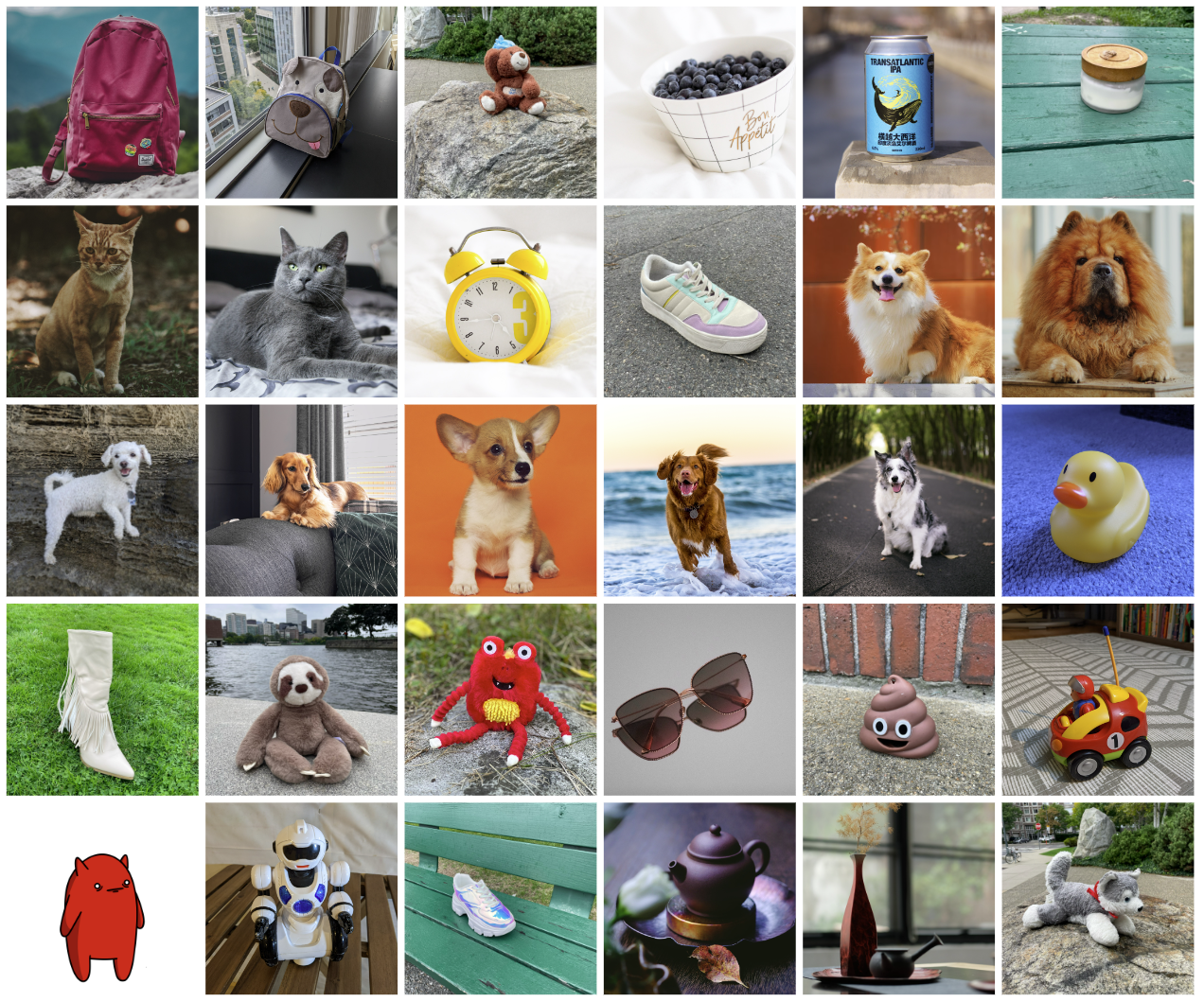}
\caption[]{
\textbf{Dataset}. Example images for each subject in our proposed dataset.
\label{fig:dataset}}
\end{figure*}

\begin{table}[t]
\centering
\resizebox{\columnwidth}{!}{
  \begin{tabular}{lcccc}
    \toprule
    Method & DINO $\uparrow$ & CLIP-I $\uparrow$ & CLIP-T $\uparrow$\\
    \midrule
    Real Images & 0.774 & 0.885 & N/A \\
    DreamBooth (Imagen) & \textbf{0.696} & \textbf{0.812} & \textbf{0.306} \\
    DreamBooth (Stable Diffusion) & 0.668 & 0.803 & 0.305 \\
    Textual Inversion (Stable Diffusion) & 0.569 & 0.780 & 0.255 \\
    \bottomrule
  \end{tabular}
}
\caption{Subject fidelity (DINO, CLIP-I) and prompt fidelity (CLIP-T, CLIP-T-L) quantitative metric comparison.
\label{table:comparison_experiment}}
\end{table}

\begin{table}[t]
\centering
\resizebox{\columnwidth}{!}{
  \begin{tabular}{lcc}
    \toprule
    Method & Subject Fidelity $\uparrow$ & Prompt Fidelity $\uparrow$\\
    \midrule
    DreamBooth (Stable Diffusion) & \textbf{68\%} & \textbf{81\%} \\
    Textual Inversion (Stable Diffusion) & 22\% & 12\% \\
    Undecided & 10\% & 7\% \\
    \bottomrule
  \end{tabular}
}
\caption{Subject fidelity and prompt fidelity user preference.
\label{table:user_study}}
\end{table}

\subsection{Dataset and Evaluation}

\paragraph{Dataset}
We collected a dataset of 30 subjects, including unique objects and pets such as backpacks, stuffed animals, dogs, cats, sunglasses, cartoons, etc. We separate each subject into two categories: objects and live subjects/pets. 21 of the 30 subjects are objects, and 9 are live subjects/pets. We provide one sample image for each of the subjects in Figure~\ref{fig:dataset}. Images for this dataset were collected by the authors or sourced from Unsplash~\cite{unsplash}.
We also collected 25 prompts: 20 recontextualization prompts and 5 property modification prompts for objects; 10 recontextualization, 10 accessorization, and 5 property modification prompts for live subjects/pets. The full list of prompts can be found in the supplementary material.

For the evaluation suite we generate four images per subject and per prompt, totaling 3,000 images. This allows us to robustly measure performances and generalization capabilities of a method. We make our dataset and evaluation protocol publicly available on the project webpage for future use in evaluating subject-driven generation.

\paragraph{Evaluation Metrics}
One important aspect to evaluate is subject fidelity: the preservation of subject details in generated images. For this, we compute two metrics: CLIP-I and DINO~\cite{caron2021emerging}. CLIP-I is the average pairwise cosine similarity between CLIP~\cite{radford2021learning} embeddings of generated and real images. Although this metric has been used in other work~\cite{gal2022image}, it is not constructed to distinguish between different subjects that could have highly similar text descriptions (e.g. two different yellow clocks). Our proposed DINO metric is the average pairwise cosine similarity between the ViT-S/16 DINO embeddings of generated and real images. This is our preferred metric, since, by construction and in contrast to supervised networks, DINO is not trained to ignore differences between subjects of the same class. Instead, the self-supervised training objective encourages distinction of unique features of a subject or image. The second important aspect to evaluate is prompt fidelity, measured as the average cosine similarity between prompt and image CLIP embeddings. We denote this as CLIP-T.

\begin{figure}[t]
\centering
\includegraphics[clip,width=1\columnwidth]{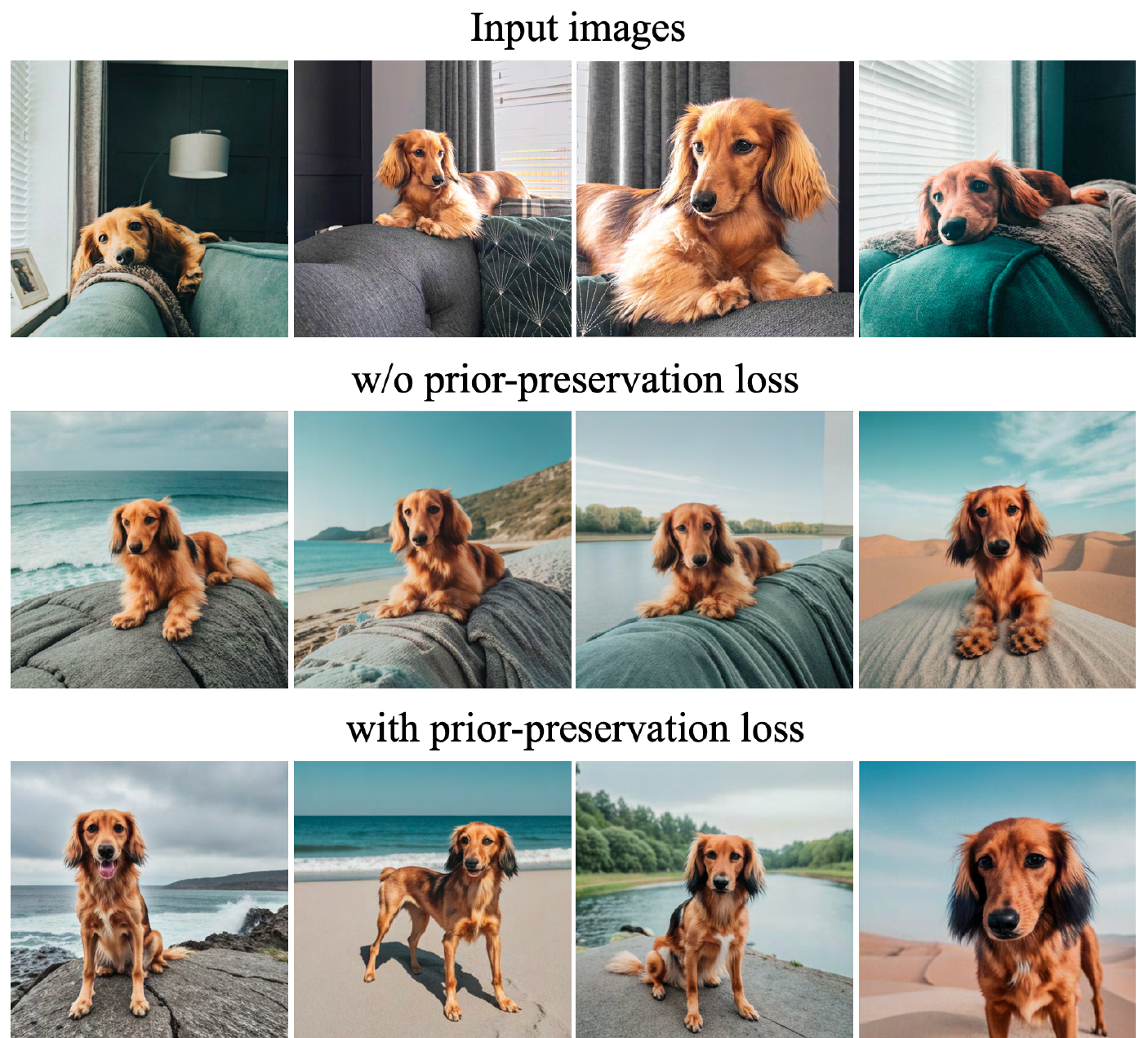}
\caption[]{
\textbf{Encouraging diversity with prior-preservation loss.} Naive fine-tuning can result in overfitting to input image context and subject appearance (e.g. pose). PPL acts as a regularizer that alleviates overfitting and encourages diversity, allowing for more pose variability and appearance diversity. 
\label{fig:prior_pres_target}
\vspace{-10pt}}
\end{figure}

\subsection{Comparisons}

We compare our results with Textual Inversion, the recent concurrent work of Gal et al.~\cite{gal2022image}, using the hyperparameters provided in their work. We find that this work is the only comparable work in the literature that is subject-driven, text-guided and generates novel images. We generate images for DreamBooth using Imagen, DreamBooth using Stable Diffusion and Textual Inversion using Stable Diffusion. We compute DINO and CLIP-I subject fidelity metrics and the CLIP-T prompt fidelity metric. In Table~\ref{table:comparison_experiment} we show sizeable gaps in both subject and prompt fidelity metrics for DreamBooth over Textual Inversion. We find that DreamBooth (Imagen) achieves higher scores for both subject and prompt fidelity than DreamBooth (Stable Diffusion), approaching the upper-bound of subject fidelity for real images. We believe that this is due to the larger expressive power and higher output quality of Imagen.

Further, we compare Textual Inversion (Stable Diffusion) and DreamBooth (Stable Diffusion) by conducting a user study. 
% We sample 300 images from each method by uniformly sampling over all 30 subjects, and randomly sample over prompts.
For subject fidelity, we asked 72 users to answer questionnaires of 25 comparative questions (3 users per questionnaire), totaling 1800 answers. Samples are randomly selected from a large pool. Each question shows the set of real images for a subject, and one generated image of that subject by each method (with a random prompt). Users are asked to answer the question: ``Which of the two images best reproduces the identity (e.g. item type and details) of the reference item?'', and we include a ``Cannot Determine / Both Equally'' option. Similarly for prompt fidelity, we ask ``Which of the two images is best described by the reference text?''. We average results using majority voting and present them in Table~\ref{table:user_study}. We find an overwhelming preference for DreamBooth for both subject fidelity and prompt fidelity. This shines a light on results in Table~\ref{table:comparison_experiment}, where DINO differences of around $0.1$ and CLIP-T differences of $0.05$ are significant in terms of user preference. Finally, we show qualitative comparisons in Figure~\ref{fig:comparison_gal}. We observe that DreamBooth better preserves subject identity, and is more faithful to prompts. We show samples of the user study in the supp. material.

\begin{figure*}[t]
\centering
\includegraphics[clip,width=1\textwidth]{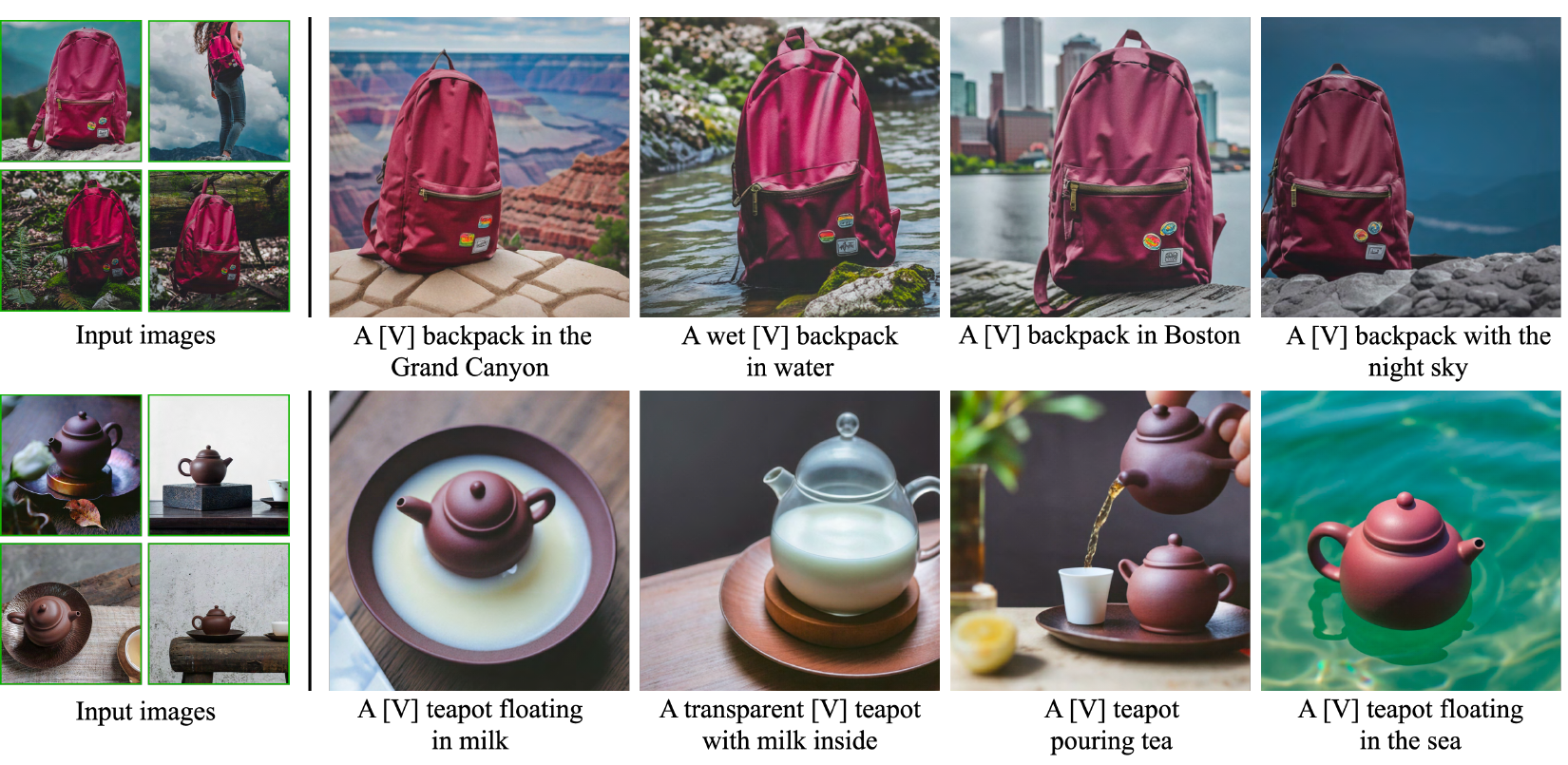}
\caption[]{
\textbf{Recontextualization.} We generate images of the subjects in different environments, with high preservation of subject details and realistic scene-subject interactions. We show the prompts below each image.
\label{fig:recontext}
\vspace{-10pt}}
\end{figure*}

\subsection{Ablation Studies}
\label{sec:app_ablation}

\paragraph{Prior Preservation Loss Ablation}
We fine-tune Imagen on 15 subjects from our dataset, with and without our proposed prior preservation loss (PPL). The prior preservation loss seeks to combat language drift and preserve the prior. We compute a prior preservation metric (PRES) by computing the average pairwise DINO embeddings between generated images of random subjects of the prior class and real images of our specific subject. The higher this metric, the more similar random subjects of the class are to our specific subject, indicating collapse of the prior. We report results in Table~\ref{table:ablation_prior_pres} and observe that PPL substantially counteracts language drift and helps retain the ability to generate diverse images of the prior class. Additionally, we compute a diversity metric (DIV) using the average LPIPS~\cite{zhang2018perceptual} cosine similarity between generated images of same subject with same prompt. We observe that our model trained with PPL achieves higher diversity (with slightly diminished subject fidelity), which can also be observed qualitatively in Figure~\ref{fig:prior_pres_target}, where our model trained with PPL overfits less to the environment of the reference images and can generate the dog in more diverse poses and articulations.

\begin{table}[t]
\centering
\resizebox{\columnwidth}{!}{
  \begin{tabular}{lcccccc}
    \toprule
    Method & PRES $\downarrow$ & DIV $\uparrow$ & DINO $\uparrow$ & CLIP-I $\uparrow$ & CLIP-T $\uparrow$  \\
    \midrule
    DreamBooth (Imagen) w/ PPL & \textbf{0.493} & \textbf{0.391} & 0.684 & 0.815 & \textbf{0.308} \\
    DreamBooth (Imagen) & 0.664 & 0.371 & \textbf{0.712} & \textbf{0.828} & 0.306 \\
    \bottomrule
  \end{tabular}
}
\caption{Prior preservation loss (PPL) ablation displaying a prior preservation (PRES) metric, diversity metric (DIV) and subject and prompt fidelity metrics.
\label{table:ablation_prior_pres}}
\end{table}

\begin{table}[t]
\centering
\resizebox{0.5\columnwidth}{!}{
  \begin{tabular}{lccccc}
    \toprule
    Method & DINO $\uparrow$ & CLIP-I $\uparrow$  \\
    \midrule
    Correct Class & \textbf{0.744} & \textbf{0.853} \\
    No Class & 0.303 & 0.607 \\
    Wrong Class & 0.454 & 0.728 \\
    \bottomrule
  \end{tabular}
}
\caption{Class name ablation with subject fidelity metrics.
\label{table:class_name}
\vspace{-12pt}}
\end{table}

\paragraph{Class-Prior Ablation}
We finetune Imagen on a subset of our dataset subjects (5 subjects) with no class noun, a randomly sampled incorrect class noun, and the correct class noun. With the correct class noun for our subject, we are able to faithfully fit to the subject, take advantage of the class prior, allowing us to generate our subject in various contexts. When an incorrect class noun (e.g. ``can'' for a backpack) is used, we run into contention between our subject and and the class prior - sometimes obtaining cylindrical backpacks, or otherwise misshapen subjects. If we train with no class noun, the model does not leverage the class prior, has difficulty learning the subject and converging, and can generate erroneous samples. Subject fidelity results are shown in Table~\ref{table:class_name}, with substantially higher subject fidelity for our proposed approach.

\subsection{Applications}

\paragraph{Recontextualization}
We can generate novel images for a specific subject in different contexts (Figure~\ref{fig:recontext}) with descriptive prompts (``a [V] [class noun] [context description]''). Importantly, we are able to generate the subject in new poses and articulations, with previously unseen scene structure and realistic integration of the subject in the scene (e.g. contact, shadows, reflections). 

\vspace{-0.2cm}

\begin{figure}[t]
\centering
\includegraphics[clip,width=1\columnwidth]{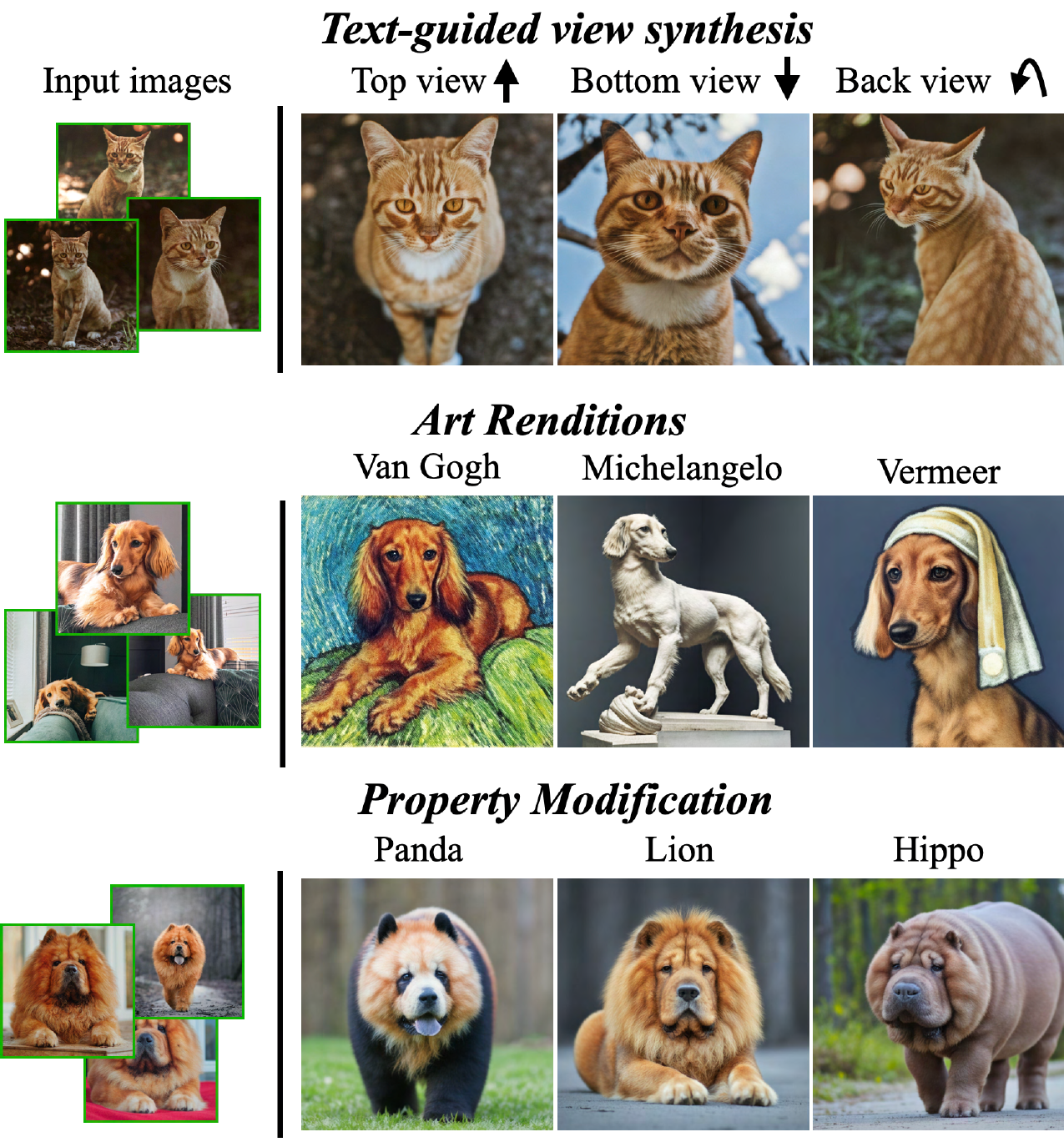}
\caption[]{
\textbf{Novel view synthesis, art renditions, and property modifications}. We are able to generate novel and meaningful images while faithfully preserving subject identity and essence. More applications and examples in the supplementary material.
\label{fig:applications}}
\end{figure}

\paragraph{Art Renditions}
Given a prompt ``a painting of a [V] [class noun] in the style of [famous painter]'' or ``a statue of a [V] [class noun] in the style of [famous sculptor]'' we are able to generate artistic renditions of our subject. Unlike style transfer, where the source structure is preserved and only the style is transferred, we are able to generate meaningful, novel variations depending on the artistic style, while preserving subject identity. E.g, as shown in Figure~\ref{fig:applications}, ``Michelangelo'', we generated a pose that is novel and not seen in the input images.

\vspace{-0.2cm}

\begin{figure}[t]
\centering
\includegraphics[clip,width=\columnwidth]{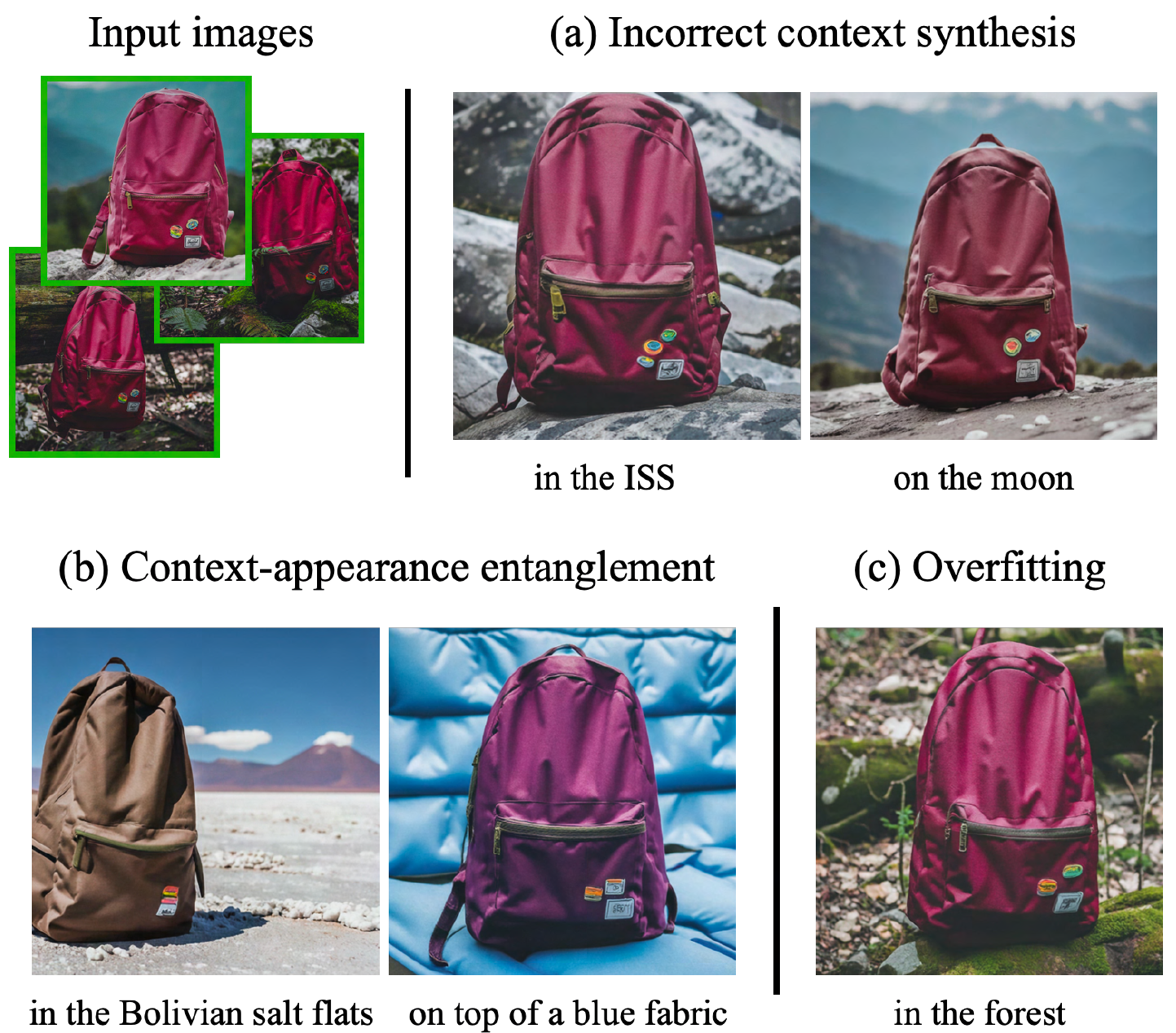}
\caption[]{
{\bf Failure modes.} Given a rare prompted context the model might fail at generating the correct environment (a). It is possible for context and subject appearance to become entangled (b). Finally, it is possible for the model to overfit and generate images similar to the training set, especially if prompts reflect the original environment of the training set (c).
\label{fig:limitations}}
\end{figure}

\paragraph{Novel View Synthesis}
We are able to render the subject under novel viewpoints. In Figure~\ref{fig:applications}, we generate new images of the input cat (with consistent complex fur patterns) under new viewpoints. We highlight that the model has not seen this specific cat from behind, below, or above - yet it is able to extrapolate knowledge from the class prior to generate these novel views given only 4 frontal images of the subject.

\paragraph{Property Modification}
We are able to modify subject properties. For example, we show crosses between a specific Chow Chow dog and different animal species in the bottom row of Figure~\ref{fig:applications}. We prompt the model with sentences of the following structure: ``a cross of a [V] dog and a [target species]''. In particular, we can see in this example that the identity of the dog is well preserved even when the species changes - the face of the dog has certain unique features that are well preserved and melded with the target species. Other property modifications are possible, such as material modification (e.g. ``a transparent [V] teapot'' in Figure~\ref{fig:recontext}). Some are harder than others and depend on the prior of the base generation model.

\subsection{Limitations}
We illustrate some failure models of our method in Figure~\ref{fig:limitations}. The first is related to not being able to accurately generate the prompted context. Possible reasons are a weak prior for these contexts, or difficulty in generating both the subject and specified concept together due to low probability of co-occurrence in the training set. The second is context-appearance entanglement, where the appearance of the subject changes due to the prompted context, exemplified in Figure~\ref{fig:limitations} with color changes of the backpack. Third, we also observe overfitting to the real images that happen when the prompt is similar to the original setting in which the subject was seen.

Other limitations are that some subjects are easier to learn than others (e.g. dogs and cats). Occasionally, with subjects that are rarer, the model is unable to support as many subject variations. Finally, there is also variability in the fidelity of the subject and some generated images might contain hallucinated subject features, depending on the strength of the model prior, and the complexity of the semantic modification.

\section{Conclusions}
We presented an approach for synthesizing novel renditions of a subject using a few images of the subject and the guidance of a text prompt. Our key idea is to embed a given subject instance in the output domain of a text-to-image diffusion model by binding the subject to a unique identifier.
Remarkably - this fine-tuning process can work given only 3-5 subject images, making the technique particularly accessible. 
We demonstrated a variety of applications with animals and objects in generated photorealistic scenes, in most cases indistinguishable from real images.

\section{Acknowledgement}
We thank Rinon Gal, Adi Zicher, Ron Mokady, Bill Freeman, Dilip Krishnan, Huiwen Chang and Daniel Cohen-Or for their valuable inputs that helped improve this work, and to Mohammad Norouzi, Chitwan Saharia and William Chan for providing us with their support and the pretrained Imagen models. Finally, a special thanks to David Salesin for his feedback, advice and for his support for the project.

% \clearpage

%%%%%%%%% REFERENCES
{\small
\bibliographystyle{ieee_fullname}
\bibliography{egbib}

\begin{thebibliography}{10}\itemsep=-1pt

\bibitem{unsplash}
Unsplash.
\newblock \url{https://unsplash.com/}.

\bibitem{abdal2021clip2stylegan}
Rameen Abdal, Peihao Zhu, John Femiani, Niloy~J Mitra, and Peter Wonka.
\newblock Clip2stylegan: Unsupervised extraction of stylegan edit directions.
\newblock {\em arXiv preprint arXiv:2112.05219}, 2021.

\bibitem{avrahami2022blendedlatent}
Omri Avrahami, Ohad Fried, and Dani Lischinski.
\newblock Blended latent diffusion.
\newblock {\em arXiv preprint arXiv:2206.02779}, 2022.

\bibitem{avrahami2022blended}
Omri Avrahami, Dani Lischinski, and Ohad Fried.
\newblock Blended diffusion for text-driven editing of natural images.
\newblock In {\em Proceedings of the IEEE/CVF Conference on Computer Vision and
  Pattern Recognition}, pages 18208--18218, 2022.

\bibitem{bar2022text2live}
Omer Bar-Tal, Dolev Ofri-Amar, Rafail Fridman, Yoni Kasten, and Tali Dekel.
\newblock Text2live: Text-driven layered image and video editing.
\newblock {\em arXiv preprint arXiv:2204.02491}, 2022.

\bibitem{Barron_2021_ICCV}
Jonathan~T. Barron, Ben Mildenhall, Matthew Tancik, Peter Hedman, Ricardo
  Martin-Brualla, and Pratul~P. Srinivasan.
\newblock Mip-nerf: A multiscale representation for anti-aliasing neural
  radiance fields.
\newblock In {\em Proceedings of the IEEE/CVF International Conference on
  Computer Vision (ICCV)}, pages 5855--5864, October 2021.

\bibitem{bau2021paint}
David Bau, Alex Andonian, Audrey Cui, YeonHwan Park, Ali Jahanian, Aude Oliva,
  and Antonio Torralba.
\newblock Paint by word, 2021.

\bibitem{boss2022samurai}
Mark Boss, Andreas Engelhardt, Abhishek Kar, Yuanzhen Li, Deqing Sun,
  Jonathan~T Barron, Hendrik Lensch, and Varun Jampani.
\newblock Samurai: Shape and material from unconstrained real-world arbitrary
  image collections.
\newblock {\em arXiv preprint arXiv:2205.15768}, 2022.

\bibitem{brock2018large}
Andrew Brock, Jeff Donahue, and Karen Simonyan.
\newblock Large scale gan training for high fidelity natural image synthesis.
\newblock {\em arXiv preprint arXiv:1809.11096}, 2018.

\bibitem{caron2021emerging}
Mathilde Caron, Hugo Touvron, Ishan Misra, Herv{\'e} J{\'e}gou, Julien Mairal,
  Piotr Bojanowski, and Armand Joulin.
\newblock Emerging properties in self-supervised vision transformers.
\newblock In {\em Proceedings of the IEEE/CVF International Conference on
  Computer Vision}, pages 9650--9660, 2021.

\bibitem{casanova2021instance}
Arantxa Casanova, Marlene Careil, Jakob Verbeek, Michal Drozdzal, and Adriana
  Romero~Soriano.
\newblock Instance-conditioned gan.
\newblock {\em Advances in Neural Information Processing Systems},
  34:27517--27529, 2021.

\bibitem{choi2021ilvr}
Jooyoung Choi, Sungwon Kim, Yonghyun Jeong, Youngjune Gwon, and Sungroh Yoon.
\newblock Ilvr: Conditioning method for denoising diffusion probabilistic
  models.
\newblock {\em arXiv preprint arXiv:2108.02938}, 2021.

\bibitem{cong2020dovenet}
Wenyan Cong, Jianfu Zhang, Li Niu, Liu Liu, Zhixin Ling, Weiyuan Li, and Liqing
  Zhang.
\newblock Dovenet: Deep image harmonization via domain verification.
\newblock In {\em Proceedings of the IEEE/CVF Conference on Computer Vision and
  Pattern Recognition}, pages 8394--8403, 2020.

\bibitem{crowson2022vqgan}
Katherine Crowson, Stella Biderman, Daniel Kornis, Dashiell Stander, Eric
  Hallahan, Louis Castricato, and Edward Raff.
\newblock Vqgan-clip: Open domain image generation and editing with natural
  language guidance.
\newblock {\em arXiv preprint arXiv:2204.08583}, 2022.

\bibitem{dhariwal2021diffusion}
Prafulla Dhariwal and Alexander Nichol.
\newblock Diffusion models beat gans on image synthesis.
\newblock {\em Advances in Neural Information Processing Systems},
  34:8780--8794, 2021.

\bibitem{ding2021cogview}
Ming Ding, Zhuoyi Yang, Wenyi Hong, Wendi Zheng, Chang Zhou, Da Yin, Junyang
  Lin, Xu Zou, Zhou Shao, Hongxia Yang, et~al.
\newblock Cogview: Mastering text-to-image generation via transformers.
\newblock {\em Advances in Neural Information Processing Systems},
  34:19822--19835, 2021.

\bibitem{ding2022cogview2}
Ming Ding, Wendi Zheng, Wenyi Hong, and Jie Tang.
\newblock Cogview2: Faster and better text-to-image generation via hierarchical
  transformers.
\newblock {\em arXiv preprint arXiv:2204.14217}, 2022.

\bibitem{esser2021taming}
Patrick Esser, Robin Rombach, and Bjorn Ommer.
\newblock Taming transformers for high-resolution image synthesis.
\newblock In {\em Proceedings of the IEEE/CVF conference on computer vision and
  pattern recognition}, pages 12873--12883, 2021.

\bibitem{gafni2022make}
Oran Gafni, Adam Polyak, Oron Ashual, Shelly Sheynin, Devi Parikh, and Yaniv
  Taigman.
\newblock Make-a-scene: Scene-based text-to-image generation with human priors.
\newblock {\em arXiv preprint arXiv:2203.13131}, 2022.

\bibitem{gal2022image}
Rinon Gal, Yuval Alaluf, Yuval Atzmon, Or Patashnik, Amit~H Bermano, Gal
  Chechik, and Daniel Cohen-Or.
\newblock An image is worth one word: Personalizing text-to-image generation
  using textual inversion.
\newblock {\em arXiv preprint arXiv:2208.01618}, 2022.

\bibitem{gal2021stylegan}
Rinon Gal, Or Patashnik, Haggai Maron, Gal Chechik, and Daniel Cohen-Or.
\newblock Stylegan-nada: Clip-guided domain adaptation of image generators.
\newblock {\em arXiv preprint arXiv:2108.00946}, 2021.

\bibitem{goodfellow2014generative}
Ian Goodfellow, Jean Pouget-Abadie, Mehdi Mirza, Bing Xu, David Warde-Farley,
  Sherjil Ozair, Aaron Courville, and Yoshua Bengio.
\newblock Generative adversarial nets.
\newblock {\em Advances in neural information processing systems}, 27, 2014.

\bibitem{hertz2022prompt}
Amir Hertz, Ron Mokady, Jay Tenenbaum, Kfir Aberman, Yael Pritch, and Daniel
  Cohen-Or.
\newblock Prompt-to-prompt image editing with cross attention control.
\newblock {\em arXiv preprint arXiv:2208.01626}, 2022.

\bibitem{hinz2020semantic}
Tobias Hinz, Stefan Heinrich, and Stefan Wermter.
\newblock Semantic object accuracy for generative text-to-image synthesis.
\newblock {\em IEEE transactions on pattern analysis and machine intelligence},
  2020.

\bibitem{ho2020denoising}
Jonathan Ho, Ajay Jain, and Pieter Abbeel.
\newblock Denoising diffusion probabilistic models.
\newblock {\em Advances in Neural Information Processing Systems},
  33:6840--6851, 2020.

\bibitem{ho2022cascaded}
Jonathan Ho, Chitwan Saharia, William Chan, David~J Fleet, Mohammad Norouzi,
  and Tim Salimans.
\newblock Cascaded diffusion models for high fidelity image generation.
\newblock {\em J. Mach. Learn. Res.}, 23:47--1, 2022.

\bibitem{jain2021dreamfields}
Ajay Jain, Ben Mildenhall, Jonathan~T. Barron, Pieter Abbeel, and Ben Poole.
\newblock Zero-shot text-guided object generation with dream fields.
\newblock 2022.

\bibitem{karras2021alias}
Tero Karras, Miika Aittala, Samuli Laine, Erik H{\"a}rk{\"o}nen, Janne
  Hellsten, Jaakko Lehtinen, and Timo Aila.
\newblock Alias-free generative adversarial networks.
\newblock {\em Advances in Neural Information Processing Systems}, 34:852--863,
  2021.

\bibitem{karras2019style}
Tero Karras, Samuli Laine, and Timo Aila.
\newblock A style-based generator architecture for generative adversarial
  networks.
\newblock In {\em Proceedings of the IEEE conference on computer vision and
  pattern recognition}, pages 4401--4410, 2019.

\bibitem{karras2020analyzing}
Tero Karras, Samuli Laine, Miika Aittala, Janne Hellsten, Jaakko Lehtinen, and
  Timo Aila.
\newblock Analyzing and improving the image quality of stylegan.
\newblock In {\em Proceedings of the IEEE/CVF Conference on Computer Vision and
  Pattern Recognition}, pages 8110--8119, 2020.

\bibitem{kim2022diffusionclip}
Gwanghyun Kim, Taesung Kwon, and Jong~Chul Ye.
\newblock Diffusionclip: Text-guided diffusion models for robust image
  manipulation.
\newblock In {\em Proceedings of the IEEE/CVF Conference on Computer Vision and
  Pattern Recognition}, pages 2426--2435, 2022.

\bibitem{kudo2018sentencepiece}
Taku Kudo and John Richardson.
\newblock Sentencepiece: A simple and language independent subword tokenizer
  and detokenizer for neural text processing.
\newblock In {\em EMNLP (Demonstration)}, 2018.

\bibitem{kwon2021clipstyler}
Gihyun Kwon and Jong~Chul Ye.
\newblock Clipstyler: Image style transfer with a single text condition.
\newblock {\em arXiv preprint arXiv:2112.00374}, 2021.

\bibitem{Lee2019CounteringLD}
Jason Lee, Kyunghyun Cho, and Douwe Kiela.
\newblock Countering language drift via visual grounding.
\newblock In {\em EMNLP}, 2019.

\bibitem{li2019controllable}
Bowen Li, Xiaojuan Qi, Thomas Lukasiewicz, and Philip Torr.
\newblock Controllable text-to-image generation.
\newblock {\em Advances in Neural Information Processing Systems}, 32, 2019.

\bibitem{li2019object}
Wenbo Li, Pengchuan Zhang, Lei Zhang, Qiuyuan Huang, Xiaodong He, Siwei Lyu,
  and Jianfeng Gao.
\newblock Object-driven text-to-image synthesis via adversarial training.
\newblock In {\em Proceedings of the IEEE/CVF Conference on Computer Vision and
  Pattern Recognition}, pages 12174--12182, 2019.

\bibitem{li2020few}
Yijun Li, Richard Zhang, Jingwan Lu, and Eli Shechtman.
\newblock Few-shot image generation with elastic weight consolidation.
\newblock {\em arXiv preprint arXiv:2012.02780}, 2020.

\bibitem{lin2018st}
Chen-Hsuan Lin, Ersin Yumer, Oliver Wang, Eli Shechtman, and Simon Lucey.
\newblock St-gan: Spatial transformer generative adversarial networks for image
  compositing.
\newblock In {\em Proceedings of the IEEE Conference on Computer Vision and
  Pattern Recognition}, pages 9455--9464, 2018.

\bibitem{liu2019more}
Xihui Liu, Dong~Huk Park, Samaneh Azadi, Gong Zhang, Arman Chopikyan, Yuxiao
  Hu, Humphrey Shi, Anna Rohrbach, and Trevor Darrell.
\newblock More control for free! image synthesis with semantic diffusion
  guidance.
\newblock 2021.

\bibitem{lu2020countering}
Yuchen Lu, Soumye Singhal, Florian Strub, Aaron Courville, and Olivier
  Pietquin.
\newblock Countering language drift with seeded iterated learning.
\newblock In {\em International Conference on Machine Learning}, pages
  6437--6447. PMLR, 2020.

\bibitem{mildenhall2020nerf}
Ben Mildenhall, Pratul~P Srinivasan, Matthew Tancik, Jonathan~T Barron, Ravi
  Ramamoorthi, and Ren Ng.
\newblock Nerf: Representing scenes as neural radiance fields for view
  synthesis.
\newblock In {\em European conference on computer vision}, pages 405--421.
  Springer, 2020.

\bibitem{mo2020freeze}
Sangwoo Mo, Minsu Cho, and Jinwoo Shin.
\newblock Freeze the discriminator: a simple baseline for fine-tuning gans.
\newblock In {\em CVPR AI for Content Creation Workshop}, 2020.

\bibitem{mokady2022self}
Ron Mokady, Omer Tov, Michal Yarom, Oran Lang, Inbar Mosseri, Tali Dekel,
  Daniel Cohen-Or, and Michal Irani.
\newblock Self-distilled stylegan: Towards generation from internet photos.
\newblock In {\em Special Interest Group on Computer Graphics and Interactive
  Techniques Conference Proceedings}, pages 1--9, 2022.

\bibitem{nichol2021glide}
Alex Nichol, Prafulla Dhariwal, Aditya Ramesh, Pranav Shyam, Pamela Mishkin,
  Bob McGrew, Ilya Sutskever, and Mark Chen.
\newblock Glide: Towards photorealistic image generation and editing with
  text-guided diffusion models.
\newblock {\em arXiv preprint arXiv:2112.10741}, 2021.

\bibitem{nichol2021improved}
Alexander~Quinn Nichol and Prafulla Dhariwal.
\newblock Improved denoising diffusion probabilistic models.
\newblock In {\em International Conference on Machine Learning}, pages
  8162--8171. PMLR, 2021.

\bibitem{nitzan2022mystyle}
Yotam Nitzan, Kfir Aberman, Qiurui He, Orly Liba, Michal Yarom, Yossi
  Gandelsman, Inbar Mosseri, Yael Pritch, and Daniel Cohen-Or.
\newblock Mystyle: A personalized generative prior.
\newblock {\em arXiv preprint arXiv:2203.17272}, 2022.

\bibitem{ojha2021few}
Utkarsh Ojha, Yijun Li, Jingwan Lu, Alexei~A Efros, Yong~Jae Lee, Eli
  Shechtman, and Richard Zhang.
\newblock Few-shot image generation via cross-domain correspondence.
\newblock In {\em Proceedings of the IEEE/CVF Conference on Computer Vision and
  Pattern Recognition}, pages 10743--10752, 2021.

\bibitem{patashnik2021styleclip}
Or Patashnik, Zongze Wu, Eli Shechtman, Daniel Cohen-Or, and Dani Lischinski.
\newblock Styleclip: Text-driven manipulation of stylegan imagery.
\newblock {\em arXiv preprint arXiv:2103.17249}, 2021.

\bibitem{poole2022dreamfusion}
Ben Poole, Ajay Jain, Jonathan~T Barron, and Ben Mildenhall.
\newblock Dreamfusion: Text-to-3d using 2d diffusion.
\newblock {\em arXiv preprint arXiv:2209.14988}, 2022.

\bibitem{qiao2019learn}
Tingting Qiao, Jing Zhang, Duanqing Xu, and Dacheng Tao.
\newblock Learn, imagine and create: Text-to-image generation from prior
  knowledge.
\newblock {\em Advances in neural information processing systems}, 32, 2019.

\bibitem{qiao2019mirrorgan}
Tingting Qiao, Jing Zhang, Duanqing Xu, and Dacheng Tao.
\newblock Mirrorgan: Learning text-to-image generation by redescription.
\newblock In {\em Proceedings of the IEEE/CVF Conference on Computer Vision and
  Pattern Recognition}, pages 1505--1514, 2019.

\bibitem{radford2021learning}
Alec Radford, Jong~Wook Kim, Chris Hallacy, Aditya Ramesh, Gabriel Goh,
  Sandhini Agarwal, Girish Sastry, Amanda Askell, Pamela Mishkin, Jack Clark,
  et~al.
\newblock Learning transferable visual models from natural language
  supervision.
\newblock {\em arXiv preprint arXiv:2103.00020}, 2021.

\bibitem{raffel2020exploring}
Colin Raffel, Noam Shazeer, Adam Roberts, Katherine Lee, Sharan Narang, Michael
  Matena, Yanqi Zhou, Wei Li, Peter~J Liu, et~al.
\newblock Exploring the limits of transfer learning with a unified text-to-text
  transformer.
\newblock {\em J. Mach. Learn. Res.}, 21(140):1--67, 2020.

\bibitem{ramesh2022hierarchical}
Aditya Ramesh, Prafulla Dhariwal, Alex Nichol, Casey Chu, and Mark Chen.
\newblock Hierarchical text-conditional image generation with clip latents.
\newblock {\em arXiv preprint arXiv:2204.06125}, 2022.

\bibitem{ramesh2021zero}
Aditya Ramesh, Mikhail Pavlov, Gabriel Goh, Scott Gray, Chelsea Voss, Alec
  Radford, Mark Chen, and Ilya Sutskever.
\newblock Zero-shot text-to-image generation.
\newblock In {\em International Conference on Machine Learning}, pages
  8821--8831. PMLR, 2021.

\bibitem{Robb2020FewShotAO}
Esther Robb, Wen-Sheng Chu, Abhishek Kumar, and Jia-Bin Huang.
\newblock Few-shot adaptation of generative adversarial networks.
\newblock {\em ArXiv}, abs/2010.11943, 2020.

\bibitem{roich2021pivotal}
Daniel Roich, Ron Mokady, Amit~H. Bermano, and Daniel Cohen-Or.
\newblock Pivotal tuning for latent-based editing of real images.
\newblock {\em ACM Transactions on Graphics (TOG)}, 2022.

\bibitem{rombach2021highresolution}
Robin Rombach, Andreas Blattmann, Dominik Lorenz, Patrick Esser, and Björn
  Ommer.
\newblock High-resolution image synthesis with latent diffusion models, 2021.

\bibitem{rombach2022high}
Robin Rombach, Andreas Blattmann, Dominik Lorenz, Patrick Esser, and Bj{\"o}rn
  Ommer.
\newblock High-resolution image synthesis with latent diffusion models.
\newblock In {\em Proceedings of the IEEE/CVF Conference on Computer Vision and
  Pattern Recognition}, pages 10684--10695, 2022.

\bibitem{palette}
Chitwan Saharia, William Chan, Huiwen Chang, Chris Lee, Jonathan Ho, Tim
  Salimans, David Fleet, and Mohammad Norouzi.
\newblock Palette: Image-to-image diffusion models.
\newblock In {\em ACM SIGGRAPH 2022 Conference Proceedings}, pages 1--10, 2022.

\bibitem{saharia2022photorealistic}
Chitwan Saharia, William Chan, Saurabh Saxena, Lala Li, Jay Whang, Emily
  Denton, Seyed Kamyar~Seyed Ghasemipour, Burcu~Karagol Ayan, S~Sara Mahdavi,
  Rapha Gontijo-Lopes, Tim Salimans, Jonathan Ho, David~J Fleet, and Mohammad
  Norouzi.
\newblock Photorealistic text-to-image diffusion models with deep language
  understanding.
\newblock {\em arXiv preprint arXiv:2205.11487}, 2022.

\bibitem{saharia2021image}
Chitwan Saharia, Jonathan Ho, William Chan, Tim Salimans, David~J Fleet, and
  Mohammad Norouzi.
\newblock Image super-resolution via iterative refinement.
\newblock {\em arXiv:2104.07636}, 2021.

\bibitem{sohl2015deep}
Jascha Sohl-Dickstein, Eric Weiss, Niru Maheswaranathan, and Surya Ganguli.
\newblock Deep unsupervised learning using nonequilibrium thermodynamics.
\newblock In {\em International Conference on Machine Learning}, pages
  2256--2265. PMLR, 2015.

\bibitem{song2020denoising}
Jiaming Song, Chenlin Meng, and Stefano Ermon.
\newblock Denoising diffusion implicit models.
\newblock In {\em International Conference on Learning Representations}, 2020.

\bibitem{song2019generative}
Yang Song and Stefano Ermon.
\newblock Generative modeling by estimating gradients of the data distribution.
\newblock {\em Advances in Neural Information Processing Systems}, 32, 2019.

\bibitem{song2020improved}
Yang Song and Stefano Ermon.
\newblock Improved techniques for training score-based generative models.
\newblock {\em Advances in neural information processing systems},
  33:12438--12448, 2020.

\bibitem{tao2020df}
Ming Tao, Hao Tang, Songsong Wu, Nicu Sebe, Xiao-Yuan Jing, Fei Wu, and Bingkun
  Bao.
\newblock Df-gan: Deep fusion generative adversarial networks for text-to-image
  synthesis.
\newblock {\em arXiv preprint arXiv:2008.05865}, 2020.

\bibitem{verbin2021refnerf}
Dor Verbin, Peter Hedman, Ben Mildenhall, Todd Zickler, Jonathan~T. Barron, and
  Pratul~P. Srinivasan.
\newblock {Ref-NeRF}: Structured view-dependent appearance for neural radiance
  fields.
\newblock {\em CVPR}, 2022.

\bibitem{Wang_2020_CVPR}
Yaxing Wang, Abel Gonzalez-Garcia, David Berga, Luis Herranz, Fahad~Shahbaz
  Khan, and Joost van~de Weijer.
\newblock Minegan: Effective knowledge transfer from gans to target domains
  with few images.
\newblock In {\em The IEEE/CVF Conference on Computer Vision and Pattern
  Recognition (CVPR)}, June 2020.

\bibitem{wu2019gp}
Huikai Wu, Shuai Zheng, Junge Zhang, and Kaiqi Huang.
\newblock Gp-gan: Towards realistic high-resolution image blending.
\newblock In {\em Proceedings of the 27th ACM international conference on
  multimedia}, pages 2487--2495, 2019.

\bibitem{xia2021tedigan}
Weihao Xia, Yujiu Yang, Jing-Hao Xue, and Baoyuan Wu.
\newblock Tedigan: Text-guided diverse face image generation and manipulation.
\newblock In {\em Proceedings of the IEEE/CVF conference on computer vision and
  pattern recognition}, pages 2256--2265, 2021.

\bibitem{yu2022scaling}
Jiahui Yu, Yuanzhong Xu, Jing~Yu Koh, Thang Luong, Gunjan Baid, Zirui Wang,
  Vijay Vasudevan, Alexander Ku, Yinfei Yang, Burcu~Karagol Ayan, et~al.
\newblock Scaling autoregressive models for content-rich text-to-image
  generation.
\newblock {\em arXiv preprint arXiv:2206.10789}, 2022.

\bibitem{zhang2018perceptual}
Richard Zhang, Phillip Isola, Alexei~A Efros, Eli Shechtman, and Oliver Wang.
\newblock The unreasonable effectiveness of deep features as a perceptual
  metric.
\newblock In {\em CVPR}, 2018.

\bibitem{zhang2018photographic}
Zizhao Zhang, Yuanpu Xie, and Lin Yang.
\newblock Photographic text-to-image synthesis with a hierarchically-nested
  adversarial network.
\newblock In {\em Proceedings of the IEEE conference on computer vision and
  pattern recognition}, pages 6199--6208, 2018.

\end{thebibliography}
}

\raggedbottom
\pagebreak
\clearpage

\section*{\LARGE Supplementary Material}

\section*{Background}
\paragraph{Text-to-Image Diffusion Models}
Diffusion models are probabilistic generative models that are trained to learn a data distribution by the gradual denoising of a variable sampled from a Gaussian distribution. Specifically, this corresponds to learning the reverse process of a fixed-length Markovian forward process. In simple terms, a conditional diffusion model $\hat\bx_\theta$ is trained using a squared error loss to denoise a variably-noised image $\bz_t \coloneqq \alpha_t \bx + \sigma_t \bepsilon$ as follows:
\begin{equation}
\label{eq:diffusion_supp}
    \Eb{\bx,\bc,\bepsilon,t}{w_t \|\hat\bx_\theta(\alpha_t \bx + \sigma_t \bepsilon, \bc) - \bx \|^2_2}
\end{equation}
where $\bx$ is the ground-truth image, $\bc$ is a conditioning vector (e.g., obtained from a text prompt), $\bepsilon \sim \mathcal{N}(\bzero, \bI)$ is a noise term and $\alpha_t, \sigma_t, w_t$ are terms that control the noise schedule and sample quality, and are functions of the diffusion process time $t \sim \mathcal{U}([0, 1])$.
At inference time, the diffusion model is sampled by iteratively denoising $\bz_{t_1} \sim \mathcal{N}(\bzero, \bI)$ using either the deterministic DDIM~\cite{song2020denoising} or the stochastic ancestral sampler~\cite{ho2020denoising}. Intermediate points $\bz_{t_1}, \dotsc, \bz_{t_T}$, where $1 = t_1 > \cdots > t_T = 0$, are generated, with decreasing noise levels. These points, $\hat{\bx}^t_0 \defeq \hat\bx_\theta(\bz_t, \bc)$, are functions of the $\bx$-predictions.

Recent state-of-the-art text-to-image diffusion models use cascaded diffusion models in order to generate high-resolution images from text~\cite{saharia2022photorealistic,ramesh2022hierarchical}. Specifically, \cite{saharia2022photorealistic} uses a base text-to-image model with 64x64 output resolution, and two text-conditional super-resolution (SR) models $64\times 64 \rightarrow 256\times 256$ and $256\times 256 \rightarrow 1024\times 1024$. Ramesh et al.~\cite{ramesh2022hierarchical} use a similar configuration, with unconditional SR models. A key component of high-quality sample generations from \cite{saharia2022photorealistic} is the use of noise conditioning augmentation~\cite{ho2022cascaded} for the two SR modules. This consists in corrupting the intermediate image using noise with specific strength, and then conditioning the SR model on the level of corruption. Saharia et al.~\cite{saharia2022photorealistic} select Gaussian noise as the form of augmentation.

Other recent state-of-the-art text-to-image diffusion models, such as Stable Diffusion~\cite{rombach2022high}, use a single diffusion model to generate high-resolution images. Specifically, the forward and backward diffusion processes occur in a lower-dimensional latent space and an encoder-decoder architecture is trained on a large image dataset to translate images into latent codes. At inference time, a random noise latent code goes through the backward diffusion process and the pre-trained decoder is used to generate the final image. Our method can be naturally applied to this scenario, where the U-Net (and possibly the text encoder) are trained, and the decoder is fixed.

\paragraph{Vocabulary Encoding}
The details of text-conditioning in text-to-image diffusion models are of high importance for visual quality and semantic fidelity. Ramesh et al.~\cite{ramesh2022hierarchical} use CLIP text embeddings that are translated into image embeddings using a learned prior, while Saharia et al.~\cite{saharia2022photorealistic} use a pre-trained T5-XXL language model~\cite{raffel2020exploring}. In our work, we use the latter. 
Language models like T5-XXL generate embeddings of a tokenized text prompt, and vocabulary encoding is an important pre-processing step for prompt embedding. In order to transform a text prompt $\bP$ into a conditioning embedding $\bc$, the text is first tokenized using a tokenizer $f$ using a learned vocabulary. Following~\cite{saharia2022photorealistic}, we use the SentencePiece tokenizer~\cite{kudo2018sentencepiece}. After tokenizing a prompt $\bP$ using tokenizer $f$ we obtain a fixed-length vector $f(\bP)$. The language model $\Gamma$ is conditioned on this token identifier vector to produce an embedding $\bc \defeq \Gamma(f(\bP))$. Finally, the text-to-image diffusion model is directly conditioned on $\bc$.

\section*{Dataset}
Our dataset includes 30 subjects. We separate each subject into two categories: objects and live subjects/pets. 21 of the 30 subjects are objects, and 9 are live subjects/pets. We provide one sample image for each of the subjects in Figure~\ref{fig:dataset}. Images for this dataset were collected by the authors or sourced from Unsplash~\cite{unsplash}.

\begin{figure*}[h!]
\centering
\includegraphics[clip,width=0.95\textwidth]{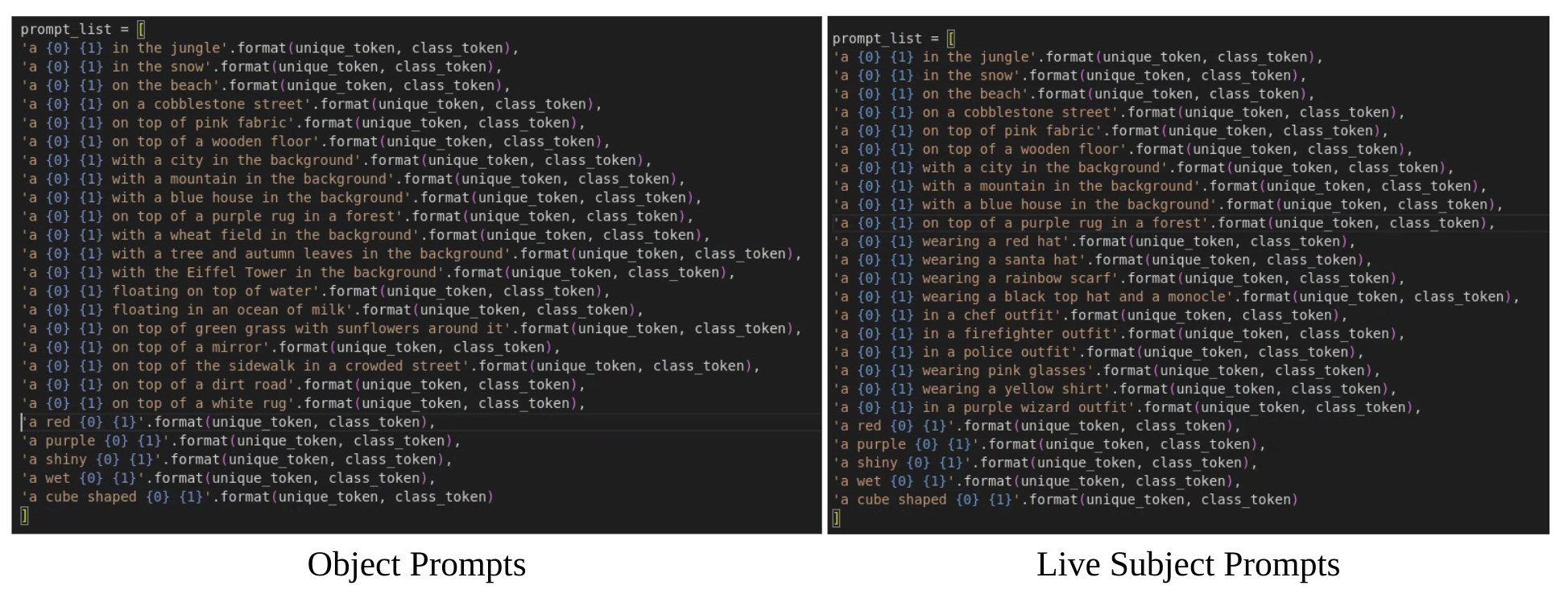}
\caption[]{
\textbf{Prompts}. Evaluation prompts for both objects and live subjects.
\label{fig:supp_prompts}}
\end{figure*}

We also collected 25 prompts: 20 recontextualization prompts and 5 property modification prompts for objects. 10 recontextualization, 10 accessorization, and 5 property modification prompts for live subjects/pets. Prompts are shown in Figure~\ref{fig:supp_prompts}

For the evaluation suite we generate four images per subject and per prompt, totaling 3,000 images. This allows us to robustly measure performances and generalization capabilities of a method. We make our dataset and evaluation protocol publicly available on the project webpage for future use in evaluating subject-driven generation.

\section*{Subject Fidelity Metrics}

\begin{figure*}[h!]
\centering
\includegraphics[clip,width=0.75\textwidth]{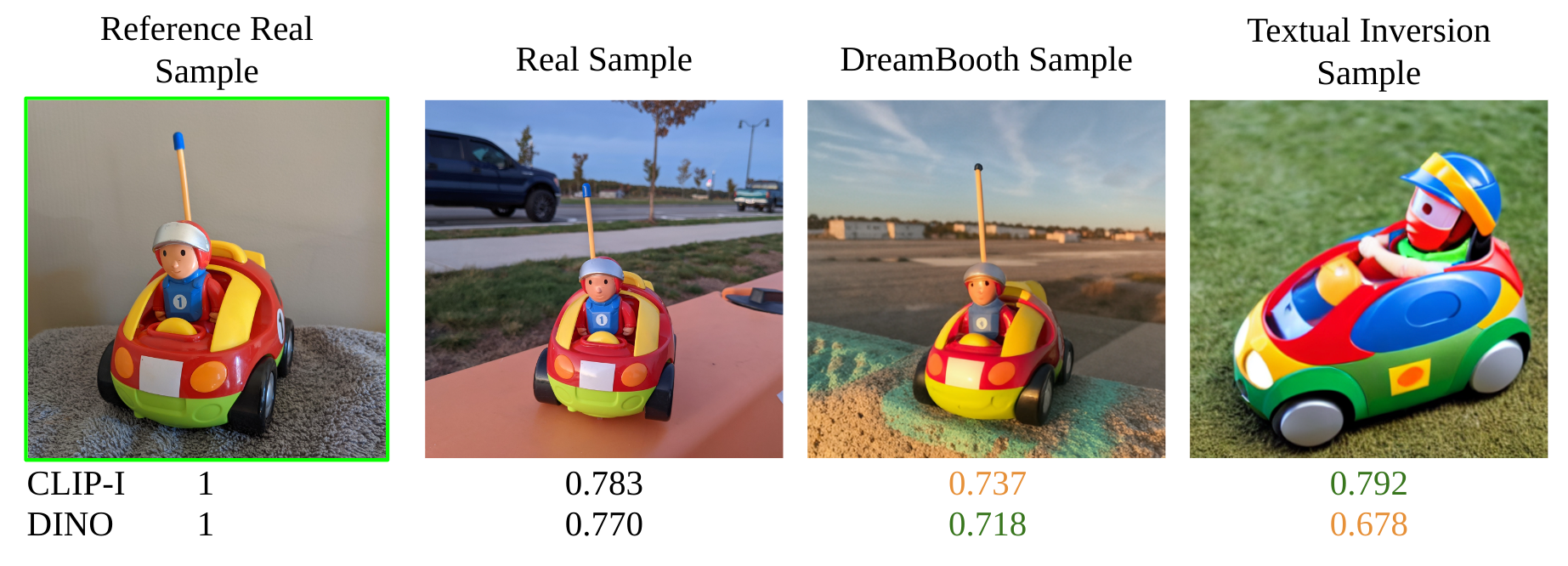}
\caption[]{
\textbf{CLIP-I vs. DINO Metrics.} The DreamBooth CLIP-I similarity to the reference image is lower than that of the Textual Inversion sample, even though the DreamBooth subject looks more similar to the reference subject. The DINO metric more closely tracks human evaluation of subject fidelity here. 
\label{fig:supp_dino_metric}}
\end{figure*}

In the main paper we comment on the superiority of our proposed DINO metric in terms of subject fidelity. We hypothesize that this is because DINO is, in essence, trained in a self-supervised manner to distinguish different images from each other modulo data augmentations. This is in contrast to the CLIP-I metric, where CLIP is trained with text-image pairs and encodes more descriptive information about images - but not necessarily fine details that are not present in the text annotations. We give an example in Figure~\ref{fig:supp_dino_metric}, where the first column contains a reference real image, the second column a different real image, the third column a DreamBooth generated image and the last column an image generated using Textual Inversion. We compare the 2nd, 3rd and 4th image to the real reference image using the CLIP-I and DINO metrics. We observe that the 2nd real image obtains both the highest CLIP-I and DINO scores. The DreamBooth sample looks much more similar to the reference sample than the Textual Inversion sample, yet the CLIP-I score for the Textual Inversion sample is much higher than the DreamBooth sample. However, we can see that the DINO similarity is higher for the DreamBooth sample - which more closely tracks human evaluation of subject fidelity. In order to quantitatively test this, we compute correlations between DINO/CLIP-I scores and normalized human preference scores. DINO has a Pearson correlation coeff.~of 0.32 with human preference (vs. 0.27 for the CLIP-I metric used in [20]), with a very low p-value of $9.44 \times 10^{-30}$.

\section*{User Study}
Below we include the full instructions used for our user study. For \textit{subject fidelity}:
\begin{itemize}
    \item Read the task carefully, inspect the reference items and then inspect the generated items.
    \item Select which of the two generated items (A or B) reproduces the identity (e.g. item type and details) of the reference item.
    \item The subject might be wearing accessories (e.g. hats, outfits). These should not affect your answer. Do not take them into account.
    \item If you're not sure, select Cannot Determine / Both Equally.
\end{itemize}
For \textit{text fidelity}:
\begin{itemize}
    \item Read the task carefully, inspect the reference text and then inspect the generated items.
    \item Select which of the two generated items (A or B) is best described by the reference text.
    \item If you're not sure, select Cannot Determine / Both Equally.
\end{itemize}
For each study we asked 72 users to answer questionnaires of 25 comparative questions (3 users per questionnaire), totaling 1800 answers - with 600 image pairs evaluated.

\section*{Additional Applications and Examples}

\paragraph{Additional Samples}
We provide a large amount of additional random samples in an annex HTML file. We compare real images, to DreamBooth generated images using Imagen and Stable Diffusion as well as images generated using Textual Inversion on Stable Diffusion.

\begin{figure*}[h!]
\centering
\includegraphics[clip,width=1\textwidth]{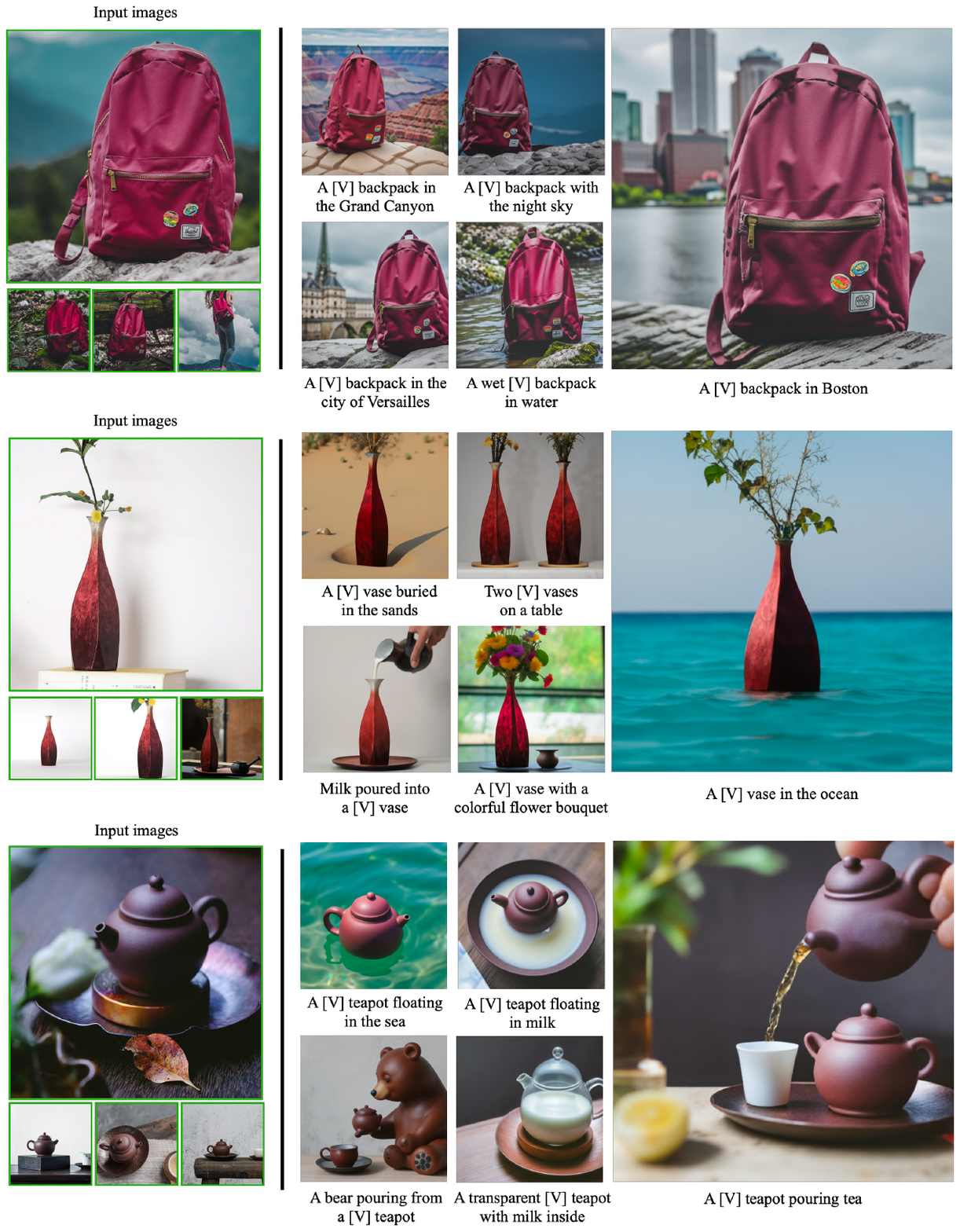}
\caption[]{
\textbf{Additional recontextualization samples of a backpack, vase, and teapot subject instances.} We are able to generate images of the subject instance in different environments, with high preservation of subject details and realistic interaction between the scene and the subject. We display the conditioning prompts below each image.\label{fig:supp_recontext}}
\end{figure*}

\paragraph{Recontextualization}
We show additional high-quality examples of recontextualization in Figure~\ref{fig:supp_recontext}.

\begin{figure*}[h!]
\centering
\includegraphics[clip,width=1\textwidth]{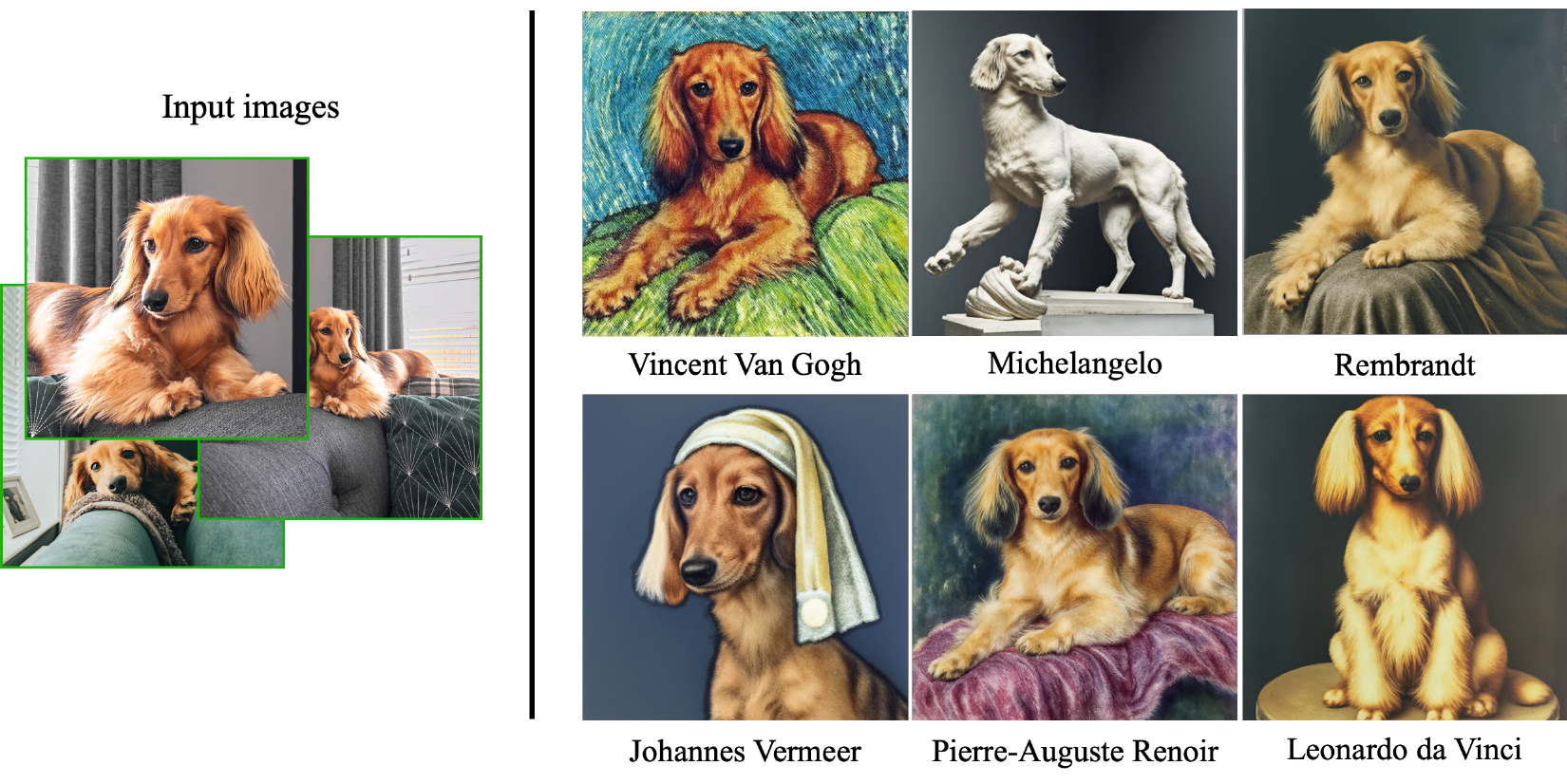}
\caption[]{
\textbf{Additional artistic renderings of a dog instance in the style of famous painters}. We remark that many of the generated poses, e.g., the Michelangelo renditions, were not seen in the training set. We also note that some renditions seem to have novel compositions and faithfully imitate the style of the painter.
\label{fig:supp_original_art}}
\end{figure*}

\paragraph{Art Renditions}
We show additional examples of original artistic renditions of a personalized model in Figure~\ref{fig:supp_original_art}.

\begin{figure*}[h!]
\centering
\includegraphics[clip,width=1\textwidth]{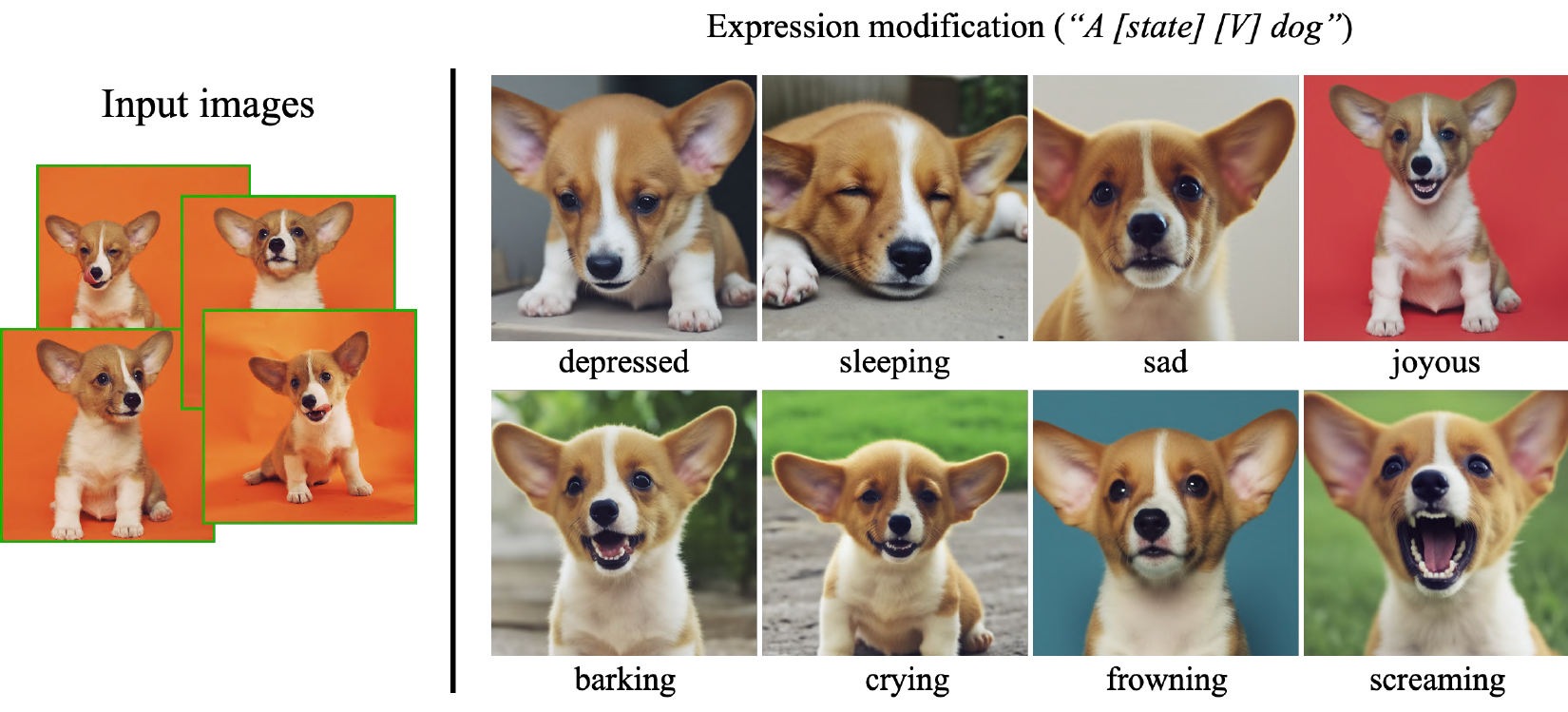}
\caption[]{
\textbf{Expression manipulation of a dog instance.} Our technique can synthesize various expressions that do not appear in the input images, demonstrating the extrapolation power of the model. Note the unique asymmetric white streak on the subject dog's face.
\label{fig:supp_expression}}
\end{figure*}

\paragraph{Expression Manipulation}
Our method allows for new image generation of the subject with modified expressions that are not seen in the original set of subject images. We show examples in Figure~\ref{fig:supp_expression}. The range of expressiveness is high, ranging from negative to positive valence emotions and different levels of arousal. In all examples, the uniqueness of the subject dog is preserved - specifically, the asymmetric white streak on its face remains in all generated images.

\begin{figure*}[h!]
\centering
\includegraphics[clip,width=1\textwidth]{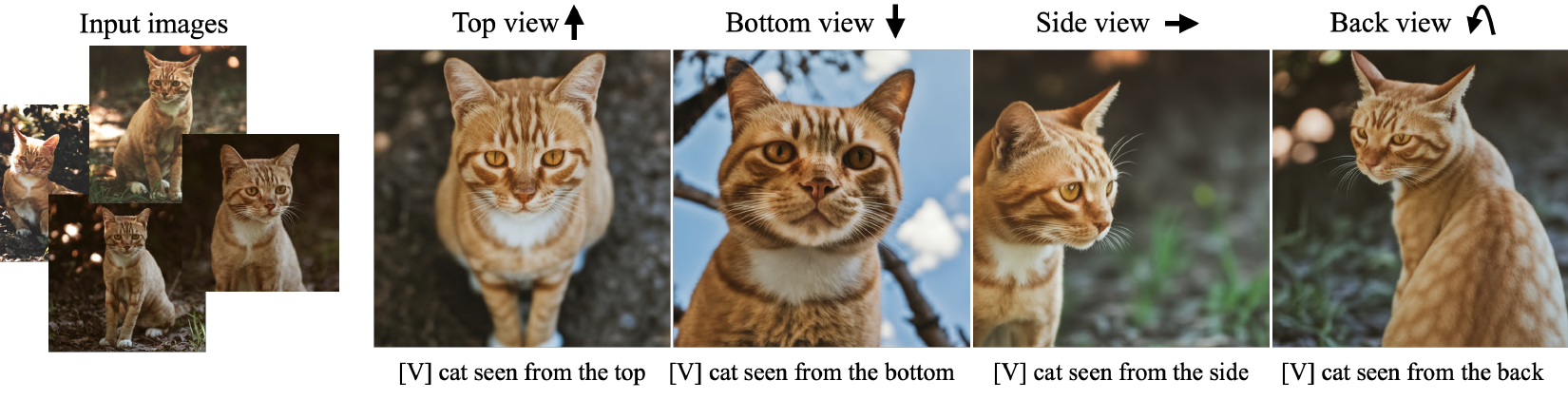}
\caption[]{
\textbf{Text-guided view synthesis}. Our technique can synthesize images with specified viewpoints for a subject cat (left to right: top, bottom, side, and back views). Note that the generated poses are different from the input poses, and the background changes in a realistic manner given a pose change. We also highlight the preservation of complex fur patterns on the subject cat's forehead.
\label{fig:supp_viewpoint}}
\end{figure*}

\paragraph{Novel View Synthesis}
We show more viewpoints for novel view synthesis in Figure~\ref{fig:supp_viewpoint}, along with prompts used to generate the samples.

\begin{figure*}[h!]
\centering
\includegraphics[clip,width=1\textwidth]{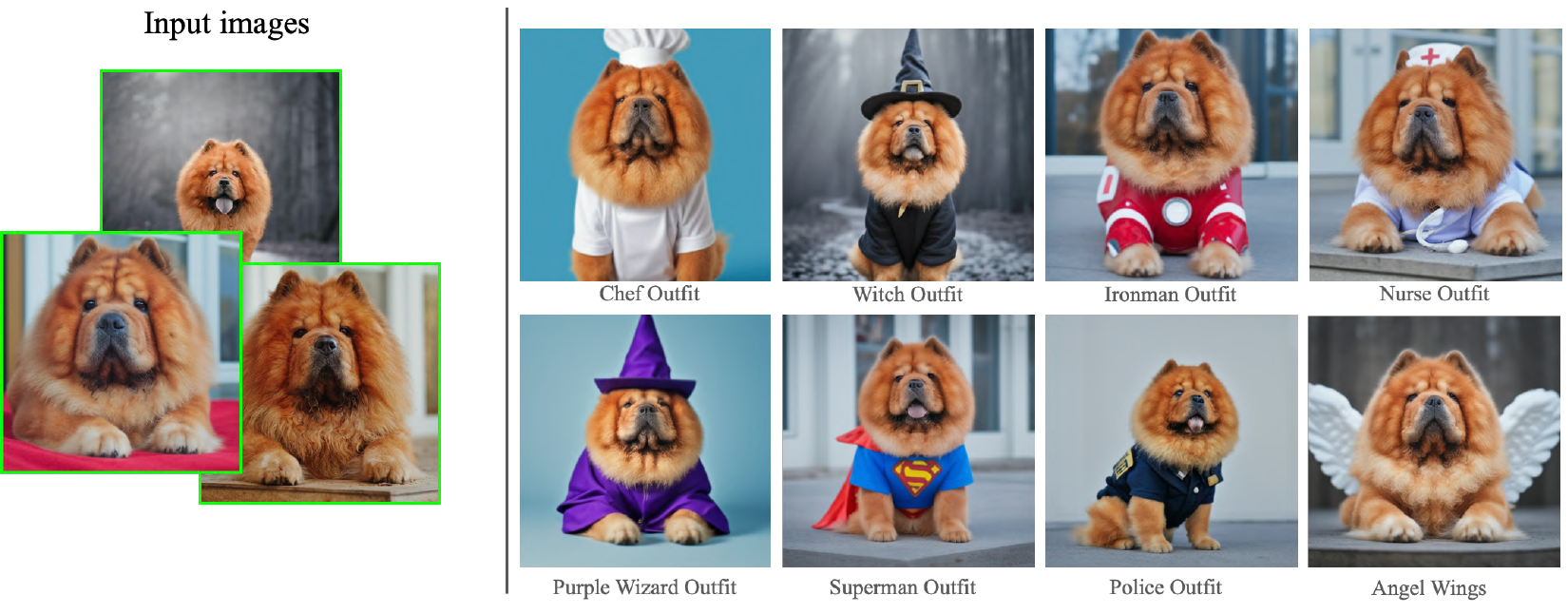}
\caption[]{
\textbf{Outfitting a dog with accessories}. The identity of the subject is preserved and many different outfits or accessories can be applied to the dog given a prompt of type ``a [V] dog wearing a police/chef/witch outfit''. We observe a realistic interaction between the subject dog and the outfits or accessories, as well as a large variety of possible options.
\label{fig:supp_accesories}}
\end{figure*}

\paragraph{Accessorization}
An interesting capability stemming from the strong compositional prior of the generation model is the ability to accessorize subjects. In Figure~\ref{fig:supp_accesories} we show examples of accessorization of a Chow Chow dog. We prompt the model with a sentence of the form: ``a [V] [class noun] wearing [accessory]''. In this manner, we are able to fit different accessories onto this dog - with aesthetically pleasing results. Note that the identity of the dog is preserved in all frames, and subject-accessory contact and articulation are realistic.

\begin{figure*}[h!]
\centering
\includegraphics[clip,width=1\textwidth]{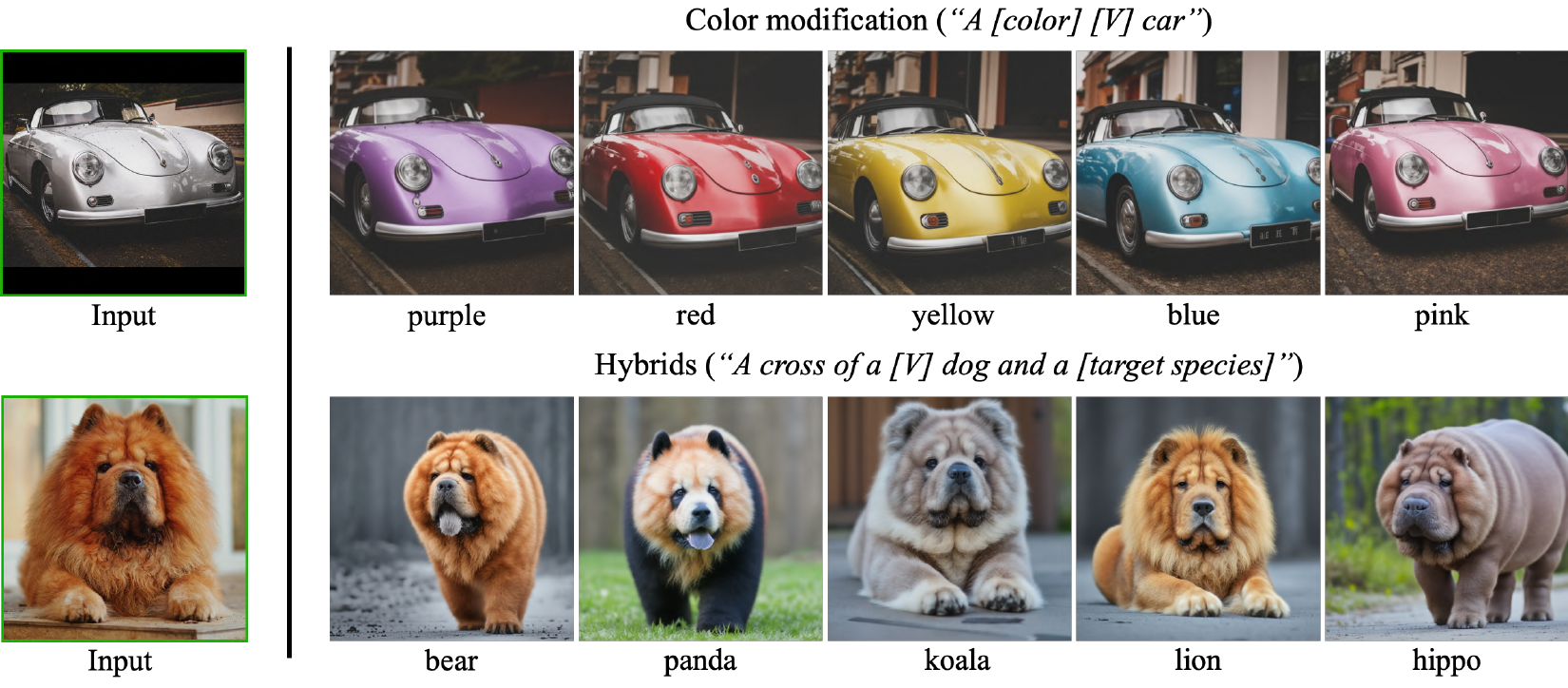}
\caption[]{
\textbf{Modification of subject properties while preserving their key features.} We show color modifications in the first row (using prompts ``a [color] [V] car''), and crosses between a specific dog and different animals in the second row (using prompts ``a cross of a [V] dog and a [target species]''). We highlight the fact that our method preserves unique visual features that give the subject its identity or essence, while performing the required property modification.
\label{fig:supp_property_mod}}
\end{figure*}

\paragraph{Property Modification}
We are able to modify subject instance properties. For example we can include a color adjective in the prompt sentence ``a [color adjective] [V] [class noun]''. In that way, we can generate novel instances of our subject with different colors. The generated scene can be very similar to the original scene, or it can be changed given a descriptive prompt. We show color changes of a car in the first row of Figure~\ref{fig:supp_property_mod}. We select similar viewpoints for effect, but we can generate different viewpoints of the car with different colors in different scenarios. This is a simple example of property modification, but more semantically complex property modifications can be achieved using our method. For example, we show crosses between a specific Chow Chow dog and different animal species in the bottom row of Figure~\ref{fig:supp_property_mod}. We prompt the model with sentences of the following structure: ``a cross of a [V] dog and a [target species]''. In particular, we can see in this example that the identity of the dog is well preserved even when the species changes - the face of the dog has certain individual properties that are well preserved and melded with the target species. Other property modifications are possible, such as material modification (e.g. a dog made out of stone). Some are harder than others and depend on the prior of the base generation model.

\begin{figure*}[h!]
\centering
\includegraphics[clip,width=\textwidth]{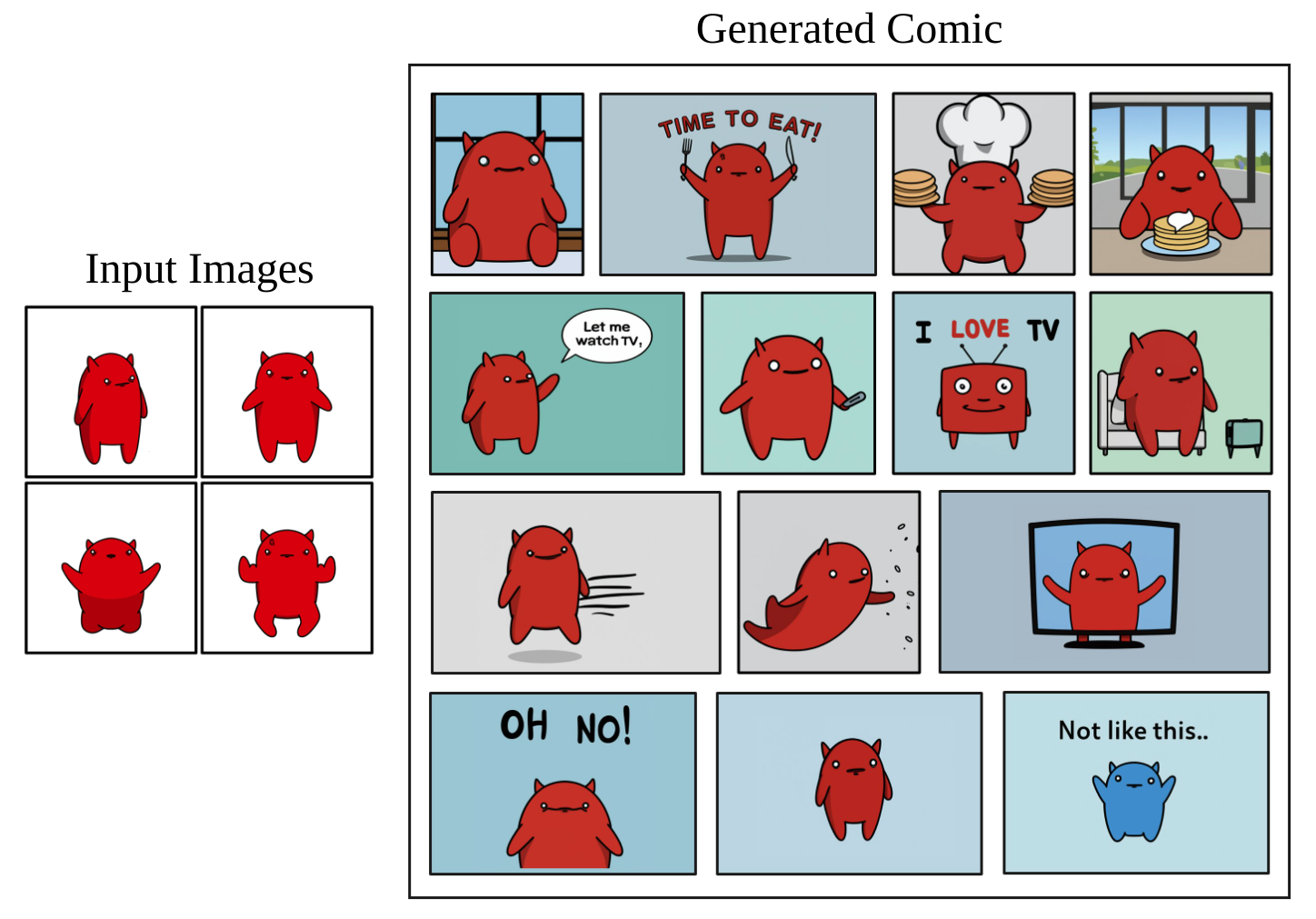}
\caption[]{\textbf{Generated comic.} We present, to the best of our knowledge, the first comic comic with a persistent character generated by prompting a generative model.
\label{fig:supp_comic}}
\end{figure*}

\paragraph{Comic Book Generation}
In addition to photorealistic images, our method is able to capture the appearance of drawn media and more. In Figure~\ref{fig:supp_comic} we present, to the best of our knowledge, the first instance of a full comic with a persistent character generated by a generative model. Each comic frame was generated using a descriptive prompt (e.g ``a [V] cartoon grabbing a fork and a knife saying ``time to eat'''').

\section*{Additional Experiments}

\subsection*{Prior Preservation Loss}
\begin{figure*}[h!]
\centering
\includegraphics[clip,width=\textwidth]{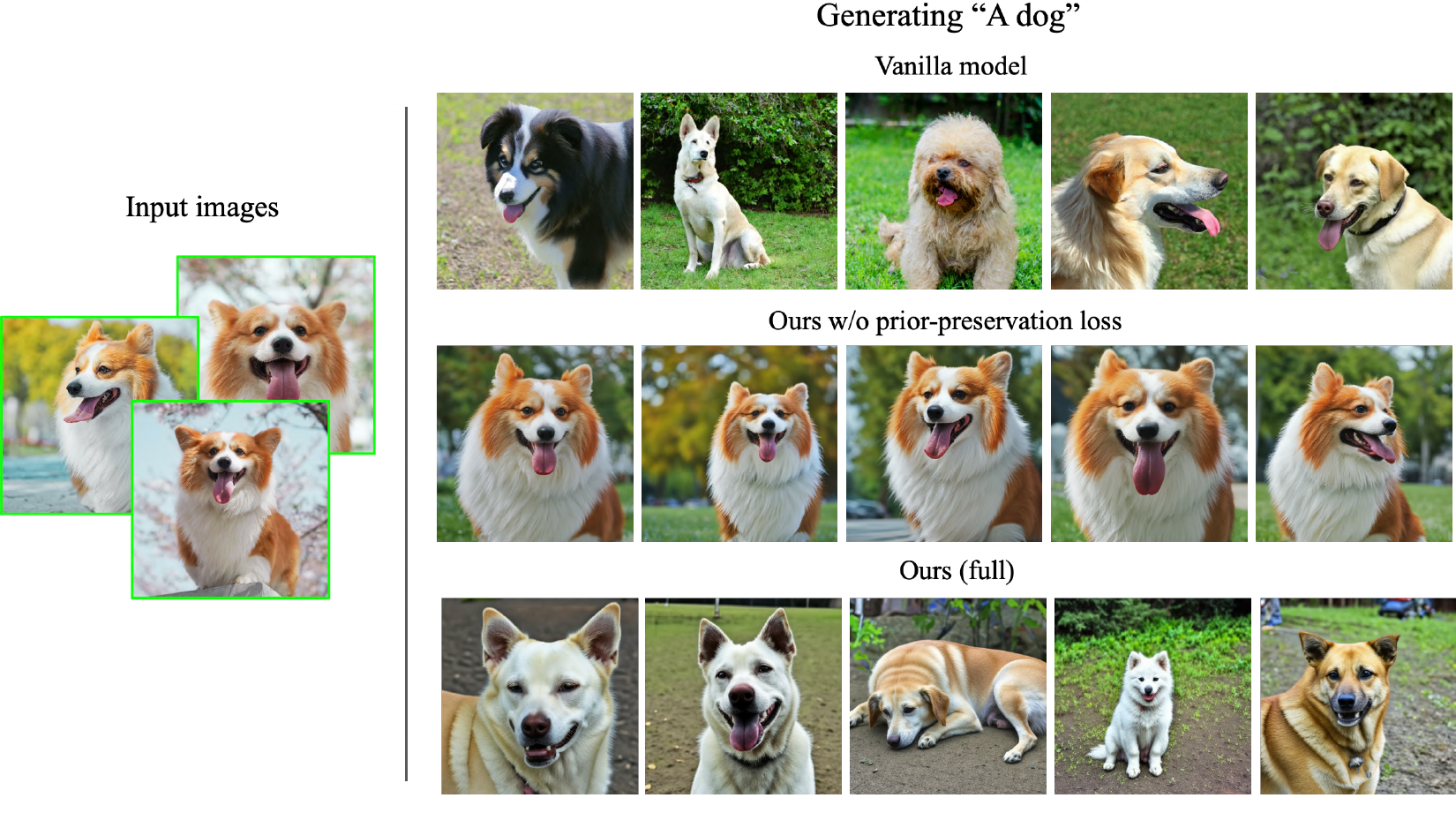}
\caption[]{\textbf{Preservation of class semantic priors with prior-preservation loss.} Fine-tuning using images of our subject without prior-preservation loss results in language drift and the model loses the capability of generating other members of our subject's class. Using a prior-preservation loss term allows our model to avoid this and to preserve the subject class' prior.
\label{fig:supp_prior_preserving}}
\end{figure*}

Here we show qualitative examples of how our prior preservation loss (PPL) conserves variability in the prior and show sample results in Figure~\ref{fig:supp_prior_preserving}. We verify that a vanilla model is able to generate a large variety of dogs, while a naively fine-tuned model on the subject dog exhibits language drift and generates our subject dog given the prompt ``a dog''. Our proposed loss preserves the variability of the prior and the model is able to generate new instances of our dog given a prompt of the style ``a [V] dog'' but also varied instances of dogs given a ``a dog'' prompt.

\subsection*{Effect of Training Images}
\begin{figure*}[h!]
\centering
\includegraphics[clip,width=\textwidth]{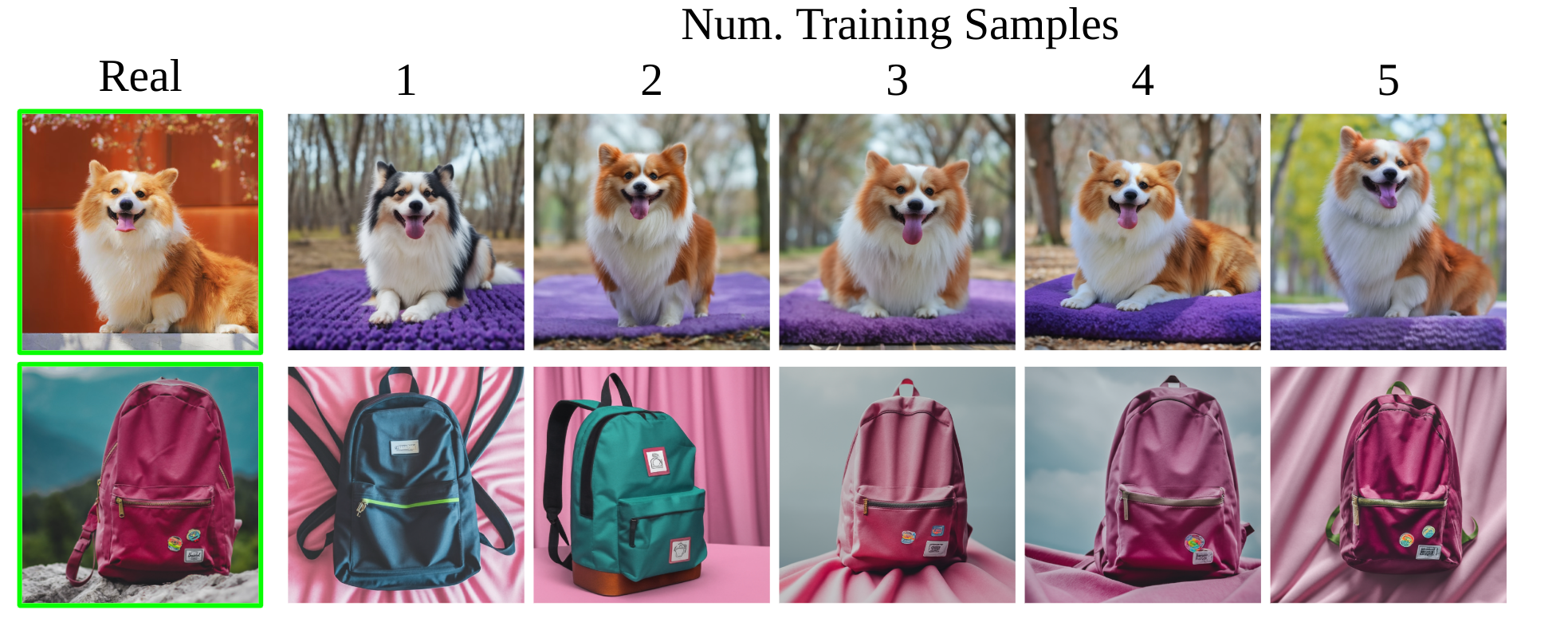}
\caption[]{\textbf{Impact of number of input images.} We observe that given only one input image, we are close to capture the identity of some subjects (e.g. Corgi dog). More images are usually needed - two images are sufficient to reconstruct the Corgi dog in this example whereas at least 3 are needed for a more rare item such as the backpack.
\label{fig:supp_number_images}}
\end{figure*}

Here we run an experiment on the effects of the number of input images for model personalization. Specifically, we train models for two subjects, 5 models per subject with input images ranging from 1 to 5. We generate 4 images for 10 different recontextualization prompts for each subject. We present qualitative results in Figure~\ref{fig:supp_number_images}. We can observe that for some subjects that are more common, and lie more strongly in the distribution of the diffusion model, such as the selected Corgi dog, we are able to accurately capture the appearance using only two images - and sometimes only one, given careful hyperparameter choice. For objects that are more rare, such as the selected backpack, we need more samples to accurately preserve the subject and to recontextualize it to diverse settings. Our quantitative results support these conclusions - we present the DINO subject fidelity metric in Table~\ref{table:number_images_dino} and the CLIP-T prompt fidelity metric in Table~\ref{table:number_images_clipt}. For both subjects we see that the optimal amount of input images for subject and prompt is 4. This number can vary depending on the subject and we settle on 3-5 images for model personalization.

\begin{table}[t]
\centering
\resizebox{0.7\columnwidth}{!}{
  \begin{tabular}{lccccc}
    \toprule
    Method & 1 & 2 & 3 & 4 & 5 \\
    \midrule
    Backpack & 0.494 & 0.515 & 0.596 & \textbf{0.604} & 0.597 \\
    Dog & 0.798 & 0.851 & 0.871 & \textbf{0.876} & 0.864 \\
    \bottomrule
  \end{tabular}
}
\caption{Effect of the number of input images on subject fidelity (DINO).
\label{table:number_images_dino}}
\end{table}

\begin{table}[t]
\centering
\resizebox{0.7\columnwidth}{!}{
  \begin{tabular}{lccccc}
    \toprule
    Method & 1 & 2 & 3 & 4 & 5 \\
    \midrule
    Backpack & 0.798 & 0.851 & 0.871 & \textbf{0.876} & 0.864 \\
    Dog & 0.646 & 0.683 & 0.734 & \textbf{0.740} & 0.730 \\
    \bottomrule
  \end{tabular}
}
\caption{Effect of the number of input images on prompt fidelity (CLIP-T).
\label{table:number_images_clipt}}
\end{table}

\subsection*{Personalized Instance-Specific Super-Resolution and Low-level Noise Augmentation for Imagen}
\begin{figure*}[h!]
\centering
\includegraphics[clip,width=\textwidth]{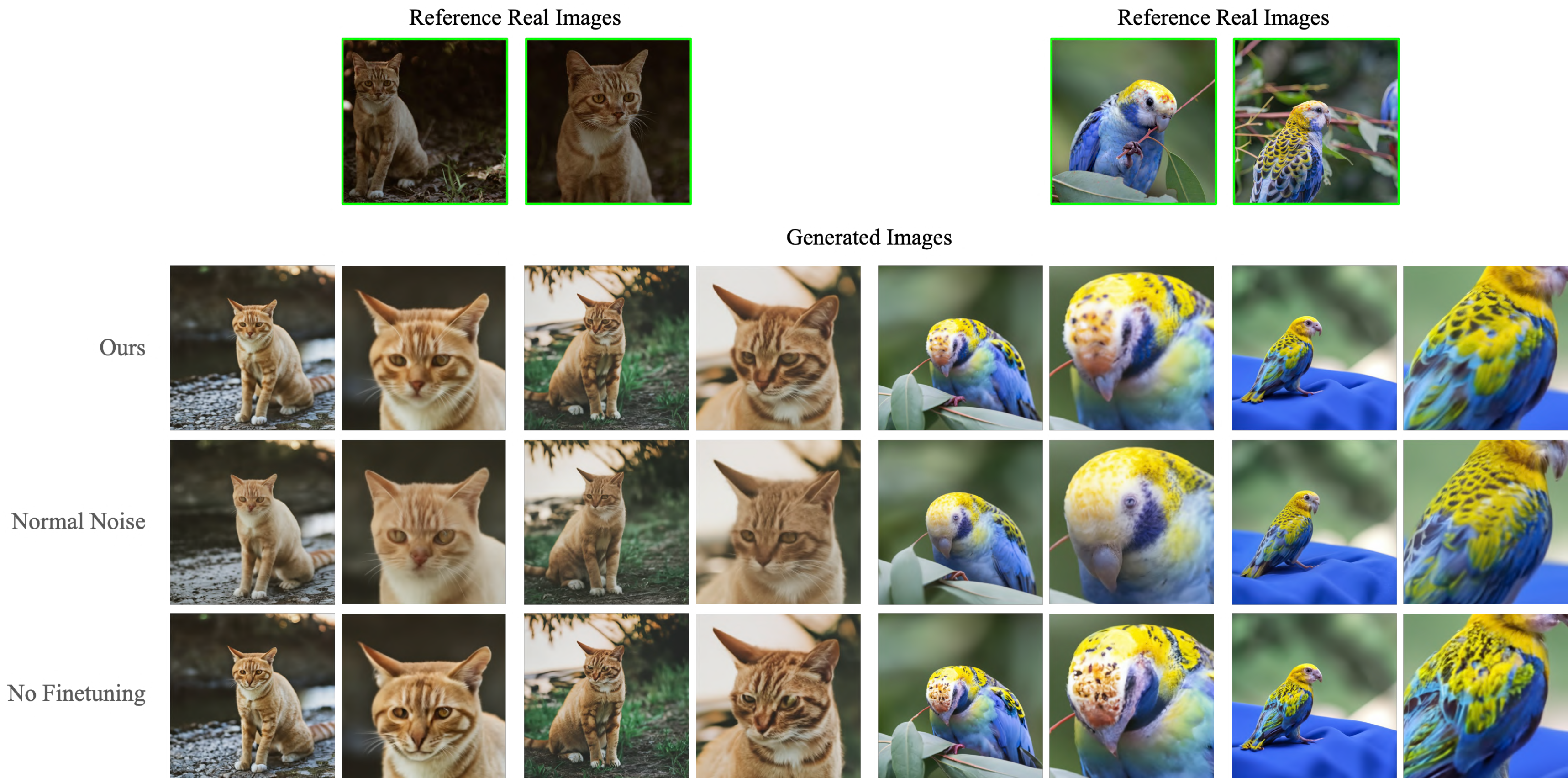}
\caption[]{
\textbf{Ablations with fine-tuning the super-resolution (SR) models.} Using the normal level of noise augmentation of \cite{saharia2022photorealistic} to train the SR models results in blurred high-frequency patterns, while no fine-tuning results in hallucinated high-frequency patterns. Using low-level noise augmentation for SR models improves sample quality and subject fidelity. Image credit (input images): Unsplash.
\label{fig:supp_low_level_noise}}
\end{figure*}

While the text-to-image diffusion model controls for most visual semantics, the super-resolution (SR) models are essential to achieve photorealistic content and to preserve subject instance details. We find that if SR networks are used without fine-tuning, the generated output can contain artifacts since the SR models might not be familiar with certain details or textures of the subject instance, or the subject instance might have hallucinated incorrect features, or missing details. Figure~\ref{fig:supp_low_level_noise} (bottom row) shows some sample output images with no fine-tuning of SR models, where the model hallucinates some high-frequency details. We find that fine-tuning the $64\times 64 \rightarrow 256\times 256$ SR model is essential for most subjects, and fine-tuning the $256\times 256 \rightarrow 1024\times 1024$ model can benefit some subject instances with high levels of fine-grained detail.

We find results to be suboptimal if the training recipes and test parameters of Saharia et al.~\cite{saharia2022photorealistic} are used to fine-tune the SR models with the given few shots of a subject instance. Specifically, we find that maintaining the original level of noise augmentation used to train the SR networks leads to the blurring of high-frequency patterns of the subject and of the environment. 
See Figure~\ref{fig:supp_low_level_noise} (middle row) for sample generations. In order to faithfully reproduce the subject instance, we reduce the level of noise augmentation from $10^{-3}$ to $10^{-5}$ during fine-tuning of the $256\times 256$ SR model. With this small modification, We are able to recover fine-grained details of the subject instance.
We show how using lower noise to train the super-resolution models improves fidelity. Specifically, we show in Figure~\ref{fig:supp_low_level_noise} that if the super-resolution models are not fine-tuned, we observe hallucination of high-frequency patterns on the subject which hurts identity preservation. Further, if we use the ground-truth noise augmentation level used for training the Imagen $256 \times 256$ model ($10^{-3}$), we obtain blurred and non-crisp details. If the noise used to train the SR model is reduced to $10^{-5}$, then we conserve a large amount of detail without pattern hallucination or blurring.

\subsection*{Comparisons}

\begin{figure*}[h!]
\centering
\includegraphics[clip,width=\textwidth]{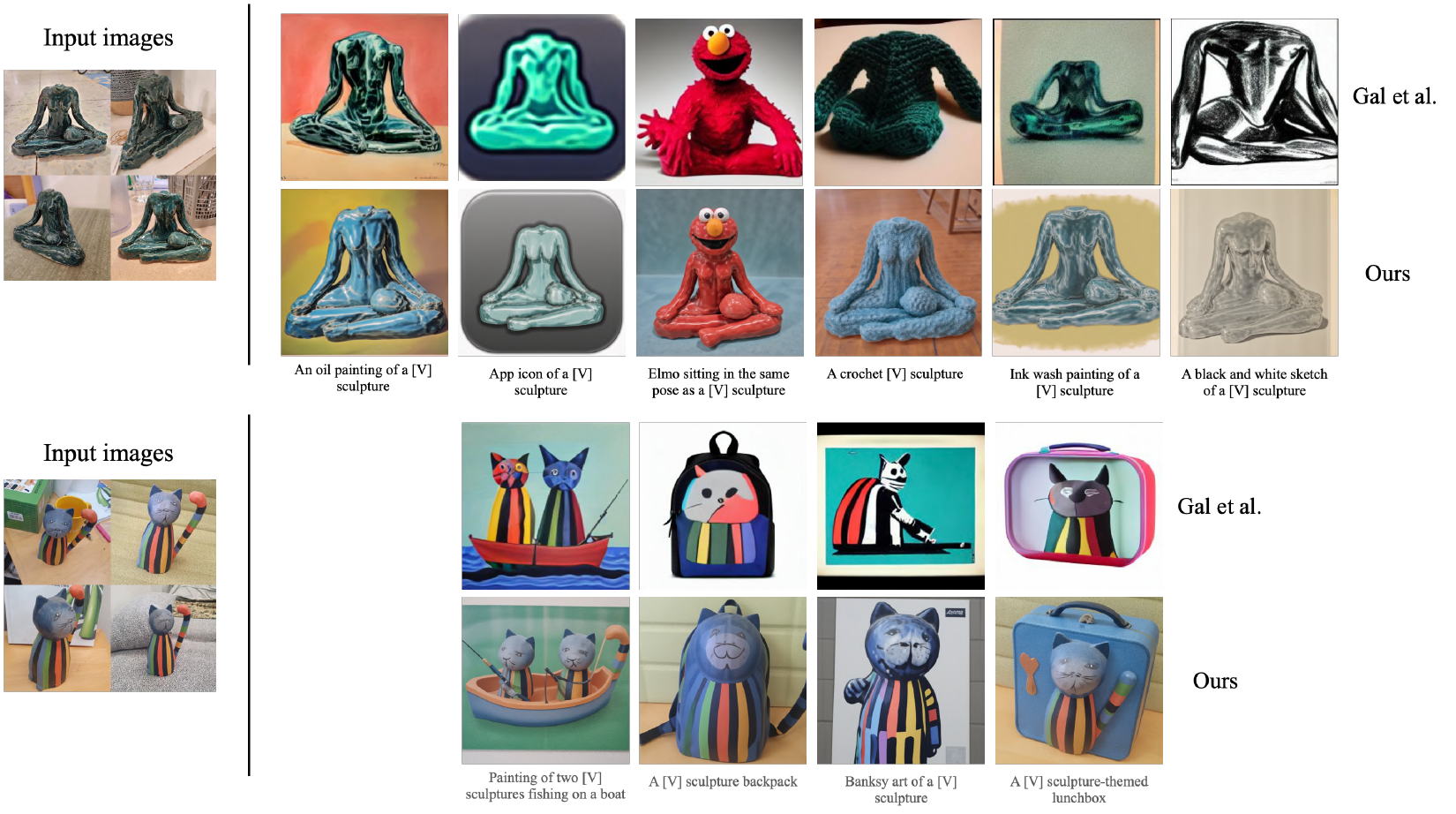}
\caption[]{
\textbf{Comparisons with Gal et al.~\cite{gal2022image}} using the subjects, images, and prompts from their work. Our approach is able to generate semantically correct variations of unique objects, exhibiting a higher degree of preservation of subject features. Input images provided by Gal et al.~\cite{gal2022image}.
\label{fig:supp_comparison_gal}}
\end{figure*}

We include additional qualitative comparisons with Gal et al.~\cite{gal2022image} in Figure~\ref{fig:supp_comparison_gal}. For this comparison, we train our model on the training images of two objects appear in the teaser of their work (headless sculpture and cat toy) kindly provided by Gal et al.~\cite{gal2022image}, and apply the prompts suggested in their paper. For prompts where they present several generated images, we handpicked their best sample (with the highest image quality and morphological similarity to the subject). We find that our work can generate the same semantic variations of these unique objects, with a high emphasis on preserving the subject identity, as can be seen, for instance, by the detailed patterns of the cat sculpture that are preserved.

\begin{figure*}[h!]
\centering
\includegraphics[clip,width=\textwidth]{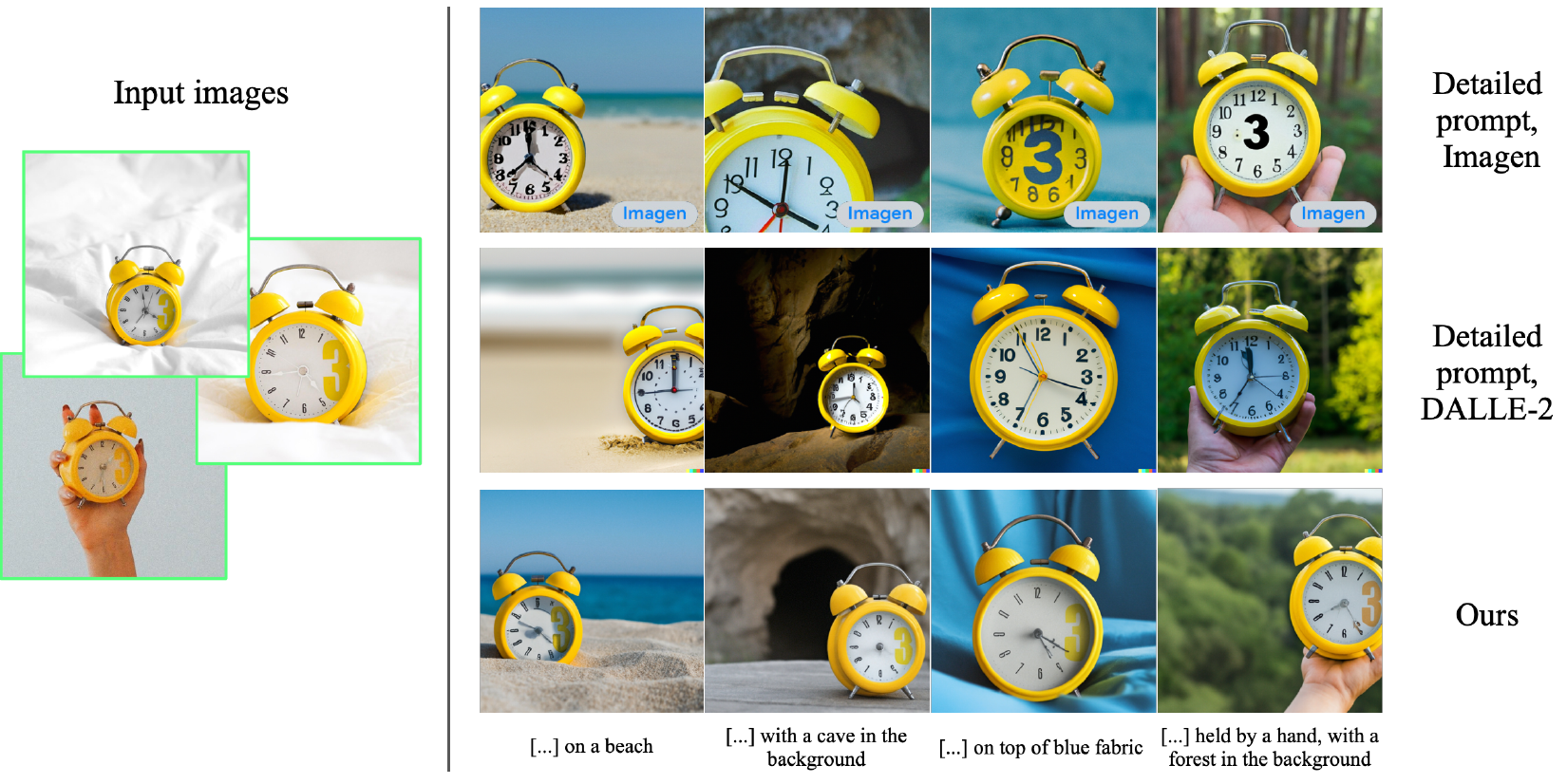}
\caption[]{
\textbf{Comparison with DALL-E 2 and Imagen with detailed prompt engineering.} After several trial-and-error iterations, the base prompt used to generate DALL-E 2 and Imagen results was \textit{``retro style yellow alarm clock with a white clock face and a yellow number three on the right part of the clock face''}, which is highly descriptive of the subject clock. In general, it is hard to control fine-grained details of subject appearance using prompts, even with large amounts of prompt engineering. Also, we can observe how context cues in the prompt can bleed into subject appearance (e.g. with a blue number 3 on the clock face when the context is ``on top of blue fabric''). Image credit (input images): Unsplash.
\label{fig:supp_comparison_prompts}}
\end{figure*}

Next, we show comparisons of recontextualization of a subject clock, with distinctive features using our method and prompt engineering using vanilla Imagen~\cite{saharia2022photorealistic} and the public API of DALL-E 2~\cite{ramesh2022hierarchical}. After multiple iterations using both models, we settle for the base prompt ``retro style yellow alarm clock with a white clock face and a yellow number three on the lower right part of the clock face'' to describe all of the important features of the subject clock example. We find that while DALL-E 2 and vanilla Imagen are able to generate retro-style yellow alarm clocks, they struggle to represent a number 3 on the clock face, distinct from the clock face numbers. In general, we find that it is very hard to control fine-grained details of subject appearance, even with exhaustive prompt engineering. Also, we find that context can bleed into the appearance of our subject instance. We show the results in Figure~\ref{fig:supp_comparison_prompts}, and can observe that our method conserves fine-grained details of the subject instance such as the shape, the clock face font, and the large yellow number three on the clock face, among others.

\section*{Societal Impact}
This project aims to provide users with an effective tool for synthesizing personal subjects (animals, objects) in different contexts. While general text-to-image models might be biased towards specific attributes when synthesizing images from text, our approach enables the user to get a better reconstruction of their desirable subjects. On contrary, malicious parties might try to use such images to mislead viewers. This is a common issue, existing in other generative models approaches or content manipulation techniques. Future research in generative modeling, and specifically of personalized generative priors, must continue investigating and revalidating these concerns.

\end{document}